\documentclass{article}

    \PassOptionsToPackage{numbers, compress}{natbib}

\usepackage[final]{main}

\usepackage[utf8]{inputenc} 
\usepackage[T1]{fontenc}    
\usepackage{hyperref}       
\usepackage{url}            
\usepackage{booktabs}       
\usepackage{amsfonts}       
\usepackage{nicefrac}       
\usepackage{microtype}      
\usepackage{xcolor}         
\usepackage{graphicx}
\usepackage{amsmath}
\usepackage{multirow} 
\usepackage{makecell}
\usepackage{caption}
\usepackage{subcaption}

\definecolor{mypink}{RGB}{239,43,159}

\title{LLaVA-UHD v2: an MLLM Integrating High-Resolution Semantic Pyramid via Hierarchical Window Transformer}

\author{
    \textbf{Yipeng Zhang}\,$^{1}$\thanks{Equal contribution.} \quad \textbf{Yifan Liu}\,$^{1*}$ \quad \textbf{Zonghao Guo}\,$^{1}$\thanks{Corresponding authors.}   \quad \textbf{Yidan Zhang}\,$^{4}$ \quad \textbf{Xuesong Yang}\,$^{4}$ \\ \quad \textbf{Xiaoying Zhang}\,$^{5}$ \quad \textbf{Chi Chen}\,$^{1}$ \quad \textbf{Jun Song}\,$^{3}$ \quad \textbf{Bo Zheng}\,$^{3}$ \quad \textbf{Yuan Yao}\,$^{2\dag}$ \\ \quad \textbf{Zhiyuan Liu}\,$^1$ 
 \quad \textbf{Tat-Seng Chua}\,$^2$ \quad \textbf{Maosong Sun\,$^1$} \\
    $^1$Tsinghua University \quad
    $^2$National University of Singapore \quad
    $^3$Alibaba Group \\
    $^4$University of Chinese Academy of Sciences \quad
    $^5$The Chinese University of Hong Kong
 \\
    \texttt{yipengzhang97@gmail.com} \quad
    \texttt{guozonghao96@outlook.com} \quad
\\ \\ 
\href{https://github.com/thunlp/LLaVA-UHD}{\textcolor{magenta}{https://github.com/thunlp/LLaVA-UHD}}
}

\begin{document}
\maketitle
\begin{abstract}
Vision transformers (ViTs) are widely employed in multimodal large language models (MLLMs) for visual encoding. 
However, they exhibit inferior performance on tasks regarding fine-grained visual perception. We attribute this to the limitations of ViTs in capturing diverse multi-modal visual levels, such as low-level details.
To address this issue, we present LLaVA-UHD v2, an MLLM with advanced perception abilities by introducing a well-designed vision-language projector, the \textbf{Hi}erarchical \textbf{win}dow (\textbf{Hiwin}) transformer. \textbf{Hiwin} transformer enhances MLLM's ability to capture diverse multi-modal visual granularities, by incorporating 
our constructed high-resolution semantic pyramid.
Specifically, \textbf{Hiwin} transformer comprises two key modules:
(i) a visual detail injection module, which progressively injects low-level visual details into high-level language-aligned semantics features, thereby forming an inverse semantic pyramid (ISP),
and 
(ii) a hierarchical window attention module, which leverages cross-scale windows to condense multi-level semantics from the ISP.
Extensive experiments show that LLaVA-UHD v2 outperforms compared MLLMs on a wide range of benchmarks.
Notably, our design achieves an average boost of 3.7\% across 14 benchmarks compared with the baseline method, 9.3\% on DocVQA for instance. All the data and code will be publicly available to facilitate future research. 
\end{abstract}

\section{Introduction}
\label{sec:intro}

\begin{figure}[!t]
\centering
\includegraphics[width=1.0\linewidth]{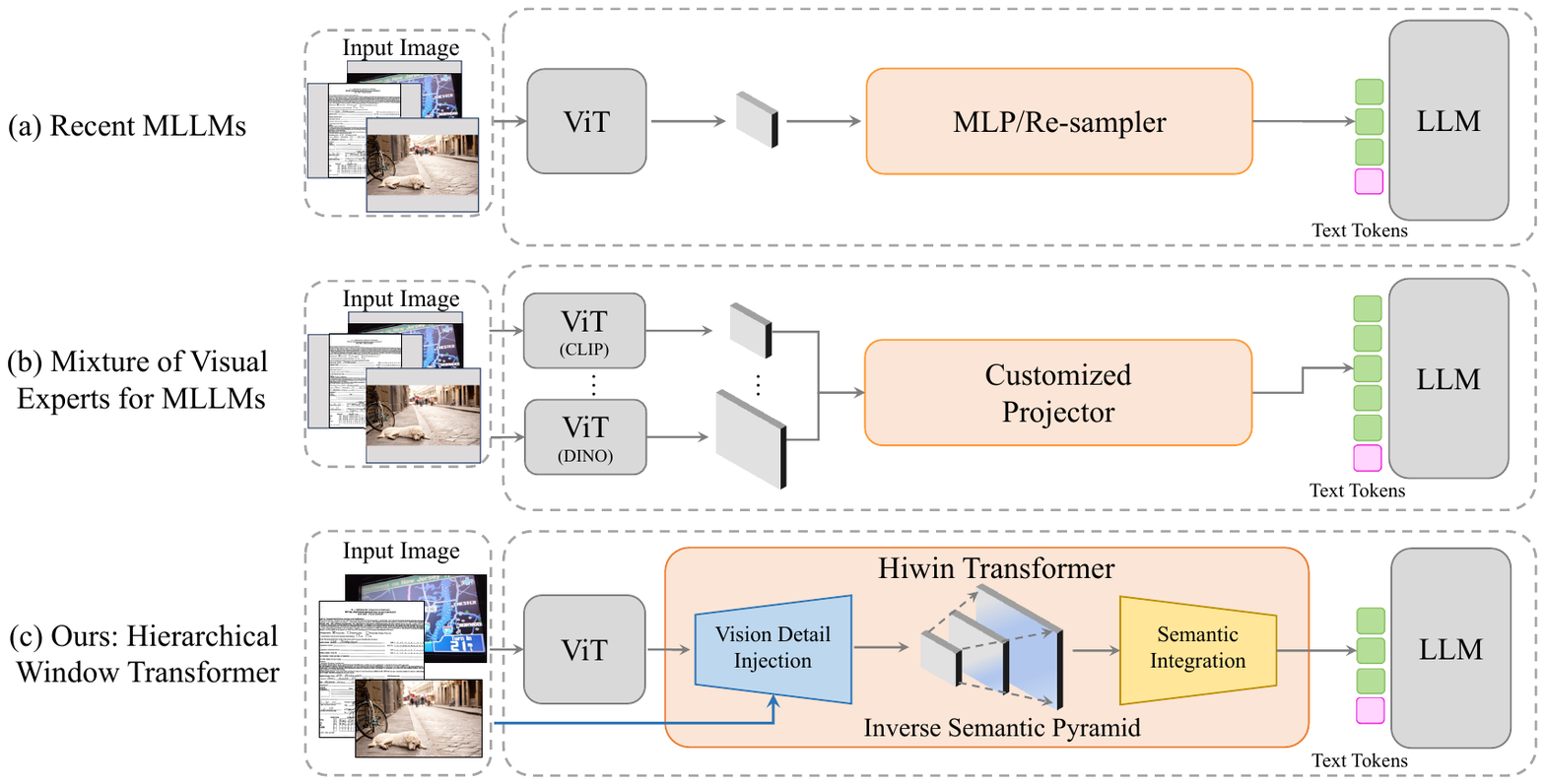}
\caption{Comparison of LLaVA-UHD v2 with other MLLMs. (a) MLLMs typically align ViT features to language space using MLPs~\cite{liu2023llava1.5} or perceiver re-samplers~\cite{Alayrac2023Flamingo,li2023blip2}, lacking visual granularity. (b) Combining multiple visual encoders is non-universal and computationally intensive. (c) LLaVA-UHD v2 employs the Hiwin transformer to build an {\color{black}inverse} semantic pyramid and compress it into visual tokens, providing various semantic granularity for language generation.
}
\vspace{-0.3cm}  
\label{fig:motivation}
\end{figure}

Embedding visual information into large language models (LLMs)~\cite{touvron2023llama,chiang2023vicuna,achiam2023gpt4,ChatGPT2022} has significantly enhanced their ability to handle multimodal tasks, such as visual question answering~\cite{antol2015vqa,hudson2019gqa,gurari2018vizwiz}, document analysis~\cite{DocVQA2021,masry2022chartqa,liu2023ocrbench}, and visual interaction~\cite{chen2024varp,gallouedec2024JAT}. 

Among these advancements, the CLIP~\cite{radford2021clip} series, built upon the vision transformer (ViT) architecture~\cite{dosovitskiy2020vit,radford2021clip}, has emerged as a standard for visual encoding in contemporary multi-modal large language models (MLLMs)~\cite{InternLM4K_2024,bai2023qwen,wang2024qwen2,liu2023llava1.5,2023llava1.6,li2023blip2,achiam2023gpt4,anil2023gemini,yao2024minicpm}. By leveraging contrastive training, CLIP-ViT extracts text-aligned visual features that facilitate seamless integration with LLMs, enabling effective handling of tasks that emphasize text-aligned semantics, $e.g.$, image captioning~\cite{plummer2015flickr30k,lin2014microsoft}.

However, CLIP-ViT-based MLLMs often underperform in tasks requiring extensive low-level visual details, such as visual grounding~\cite{yu2016rec,chen2024augrec,zhan2025griffon,you2023ferret} and optical character recognition~\cite{kim2022donut,lee2023pix2struct}, hindering the MLLMs' practical applications. As illustrated in Fig.~\ref{fig:motivation}(a), such inferior performance primarily stems from the low resolution and text-aligned nature of CLIP features. 
To facilitate MLLMs' fine-grained perception, \citet{yao2025dense} proposes a Dense Connector~\cite{yao2025dense}, which fuse lower layers of CLIP-ViT. Yet, its performance is limited by the low resolution of the features.

To address such limitation, we attempts to incorporate a vision feature pyramid~\cite{lin2017fpn,ronneberger2015unet,zhu2020deformable} into MLLMs to provide multi-level visual granularity. However, directly employing such pyramid might encounter the following two key challenges:
(1) \textit{Representation Non-inheritance.} 
The inherent image-text alignment of CLIP-ViT’s feature representations is fundamental to the efficacy of MLLMs. Substituting CLIP-ViT with hierarchical architectures such as Swin~\cite{liu2021swin} to obtain multi-scale representations would disrupt this alignment and consequently compromise the transferability of its pretrained representations.
(2) \textit{Compression Ineffectiveness.} The quadratic computational cost of LLMs w.r.t. the number of visual tokens necessitates effective compression of the feature pyramid. 
Current projecting methods~\cite{shi2024eagle,SPHINX2023,lu2024deepseekvl,tong2024cambrian} 
designed a customized projector to resize multi-scale visual features before token compression, which resulted in a loss in visual details and spatial relations in naive feature resolution, as shown in Fig.~\ref{fig:motivation}(b).

To address these issues, we present LLaVA-UHD v2, an advanced MLLM with enhanced perceptual capabilities leveraging the incorporated \textbf{Hiwin} transformer. As shown in Fig.~\ref{fig:motivation}(c), \textbf{Hiwin} transformer enables capturing diverse multi-modal granularity by integrating a constructed high-resolution semantic pyramid.

Specifically, the Hiwin transformer consists of two key modules as follows: \textbf{(i) a visual detail injection module (VDIM) for inheriting the representations.} We propose a VDIM to inject low-level details ($e.g.$, edges, textures) from images into text-aligned features from a CLIP-pretrained ViT~\cite{radford2021clip}, progressively building an up-sampled level upon the previous level, resulting in an inverse semantic pyramid (ISP). During the training stage, a reconstruction loss between fused and original CLIP features explicitly maintains vision-language alignment while enhancing visual granularity. This strategy can be extended to any ViT for inheriting its powerful multi-modal representations. \textbf{(ii) a hierarchical window attention module for ensuring effective compression}. We propose utilizing a set of hierarchical windows to capture semantics from local regions across different semantic levels. A set of learnable queries is restricted to attending solely to sampled features within their respective windows. 
This attention mechanism performs effective compression on local dense features at the native resolution of each pyramid level, thereby enabling visual tokens to effectively capture both fine-grained visual details and high-level language-aligned semantics.

Extensive experiments demonstrate that LLaVA-UHD v2 dramatically outperforms the compared MLLMs across a wide range of benchmarks. More importantly, LLaVA-UHD v2 surpasses the baseline method (LLaVA-UHD~\cite{guo2024llava-uhd}) on {\color{black}14} popular benchmarks by 3.7\% in average, including document-centric visual question answering ($e.g.$, +9.3\% on DocVQA), visual grounding ($e.g.$, average +5.7\% on RefCOCOs~\cite{yu2016rec}), and high-resolution image perception ($e.g.$, +3.4\% on HR-Bench~\cite{wang2024hrbench}). Besides, we experimentally reveal that ISP, regardless of construction method ($e.g.$, bilinear interpolation), enhances the visual perception capabilities of MLLMs, offering new insights for future research.
In summary, our contributions are three-fold:
\begin{itemize}
    \item  
        We present LLaVA-UHD v2, an MLLM with the advanced visual perception ability, through the integration of a high-resolution semantic pyramid representation.
    \item 
        We propose the Hiwin transformer, a novel vision-language projector comprises a visual detail injection module and a hierarchical window attention module for capturing diverse multimodal visual granularities.
    \item 
        Trained on merely academic-scale data, LLaVA-UHD v2 achieves substantial improvements over the baseline method across 14 benchmarks.
\end{itemize}

\section{Related Works}
\label{sec:related}

\begin{figure*}[!t]
\centering
\includegraphics[width=1.0\linewidth]{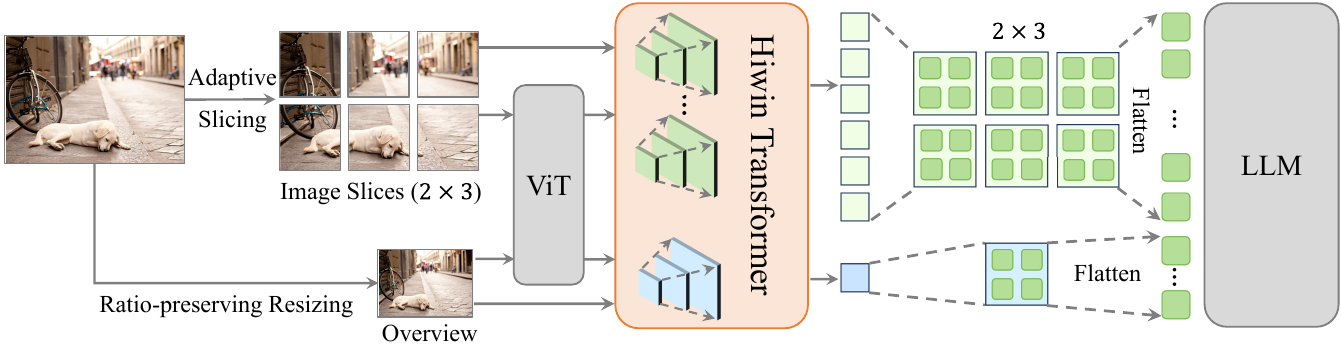}
\caption{The overall architecture of proposed LLaVA-UHD v2, consisting of a ViT, our hierarchical window transformer (Hiwin transformer), and an LLM. The Hiwin transformer first injects high-frequency visual details from the image into the high-level semantics of ViT features, forming inverse semantic pyramids (ISP). Then it compresses the ISPs into spatially consistent tokens via cross-scale windows, for a better vision-language alignment. Details about the two procedures are illustrated in Figure \ref{fig:featup_module} and \ref{fig:hiwin-attn}.}

\vspace{-0.3cm}  
\label{fig:pipe}
\end{figure*}

\subsection{Feature Pyramid Representation}

Image pyramid techniques are fundamental in image processing, facilitating multi-resolution analysis since the era of manual feature design, as seen in SIFT’s scale-space keypoints~\cite{ImagePyramid1983, sift2004}. In deep learning, CNNs such as ResNet and VGG, inherently extracts hierarchical features across scales~\cite{he2016resnet, simonyan2014vgg}. Innovations such as FPN and U-Net enhance semantic hierarchy for tasks like detection and segmentation~\cite{lin2017fpn, ronneberger2015unet}. Recently, some transformer-based models~\cite{liu2021swin,zhang2023hivit,zhu2020deformable} further advanced feature pyramid construction, capturing more comprehensive visual semantic granularity for visual representations. However, multimodal language models, often using CLIP-based ViTs, underutilize hierarchical features, suggesting a research gap for integrating advanced feature pyramids into these models~\cite{zhang2023hivit, radford2021clip}.

\subsection{Visual Encoding in MLLMs} 
CLIP-ViT, favored for its effective alignment of visual features with linguistic semantics through contrastive pre-training, is widely adopted in MLLMs~\cite{radford2021clip,liu2024llava,liu2023llava1.5,bai2023qwen,2023llava1.6,anil2023gemini,Zhu2023MiniGPT4,li2023monkey}. Emerging research explores alternative visual representations, primarily in three categories:
(1) Fusing features from CLIP-based CNNs and ViTs. LLaVA-HR~\cite{luo2024llavahr} integrates stage-wise CNN features~\cite{liu2022convnet} into ViT’s layers, enhancing the fine-grained perception of ViT representations. CogAgent~\cite{hong2023cogagent} utilizes ViT features to query high-resolution CNN features for detailed information during language decoding. Mini-Gemini~\cite{li2024minigemini} employs a cross-attention-based post-fusion between CNN and ViT features.
(2) Fusing features from visual experts trained with different vision pre-training tasks~\cite{SPHINX2023,gao2024sphinxx,lu2024deepseekvl,zong2024mova,tong2024cambrian,shi2024eagle,wei2023vary}. Candidate experts include DINO-v2~\cite{oquab2023dinov2} by visual contrastive pre-training, SAM~\cite{Kirillov2023SAM} by prompt segmentation pre-training, Pix2Struct~\cite{lee2023pix2struct} by document parsing pre-training, $etc.$ Deepseek-VL~\cite{lu2024deepseekvl}, SPHINX-X~\cite{SPHINX2023,gao2024sphinxx} and Eagle~\cite{shi2024eagle} down-sample output feature maps and concatenate them along the channel axis. Cambrian-1~\cite{tong2024cambrian} initializes embeddings to query local patches from visual experts.
(3) Language models for visual encoding: Fuyu~\cite{fuyu2023}, Otter-HD~\cite{li2023otterhd}, and SOLO~\cite{chen2024solo} encode images directly with LLMs, bypassing dedicated visual encoders. However, these approaches, while effective, increase computational demand and hinder a unified image-to-language design.
\subsection{Token Projection and Compression}
MLPs~\cite{liu2024llava,liu2023llava1.5}, perceiver resamplers~\cite{Alayrac2023Flamingo,bai2023qwen,yao2024minicpm} and Q-Formers~\cite{li2023blip2,instructblip2023,zhang2024beyondllavahd} are basic projectors widely used in modern MLLMs. 
Recently, various new designs have emerged.
(1) Spatial-preserving compression. Qwen2-VL~\cite{wang2024qwen2} and MiniGPT-v2~\cite{chen2023minigptv2} employ a simple linear to merge tokens locally ($e.g.$ 2$\times$2). Honey-bee~\cite{cha2024honeybee} introduces a C-abstractor ($i.e.$ CNN-based block), while Oryx~\cite{liu2024oryx} dynamically pools features and utilizes them to query the original ones.
(2) Cross-layer feature compression. Token-Packer~\cite{li2024tokenpacker} compresses features across layers in a ViT using cross-attention. MMFuser~\cite{cao2024mmfuser} uses final-layer features as queries to attend to early-layer features. 
(3) Semantic-merged compression. Chat-UniVi~\cite{jin2024chatunivi} and LLaVA-PruMerge~\cite{shang2024llavaprumerge} extending the token merging~\cite{bolya2022tome} strategy, merge features with similar semantics into a single representation. 
However, these approaches depend heavily on feature padding, resizing, and reshaping, hindering their ability to compress features with arbitrary resolutions.
\section{Method}
\label{sec:method}

\subsection{Overview}

The architecture of the proposed LLaVA-UHD v2 is illustrated in Fig.~\ref{fig:pipe}. 
LLaVA-UHD v2 comprises three modules: a visual encoder (ViT), a vision-language projector (Hiwin transformer), and an LLM.
We first partition and process the input image with the adaptive slicing strategy from \cite{guo2024llava-uhd}, outputting CLIP~\cite{radford2021clip} features with arbitrary size and shape.
The resulting features are subsequently passed to the Hiwin transformer for vision-language projection, which is carried out with two stages: (i) constructing an inverse semantic pyramid (ISP) and (ii) integrating the ISP by hierarchical window attention, which will be detailed in Sec.\ref{subsec:pyramid} and Sec.\ref{subsec:window}. 
The core of the Hiwin transformer lies in 
enhancing each ViT-encoded feature into a high-resolution semantic pyramid encoding, thereby achieving enriched semantic granularity for each slice or image.
After the enhancement, visual tokens from different slices are reorganized into a spatially consistent feature map relative to the original image, ensuring clarity in spatial relationships. 
Followed by concatenating with the overview tokens, the visual tokens provide both high-level language-aligned semantics and high-resolution visual details for language decoding.

\begin{figure*}[!t]
\centering
\includegraphics[width=1.0\linewidth]{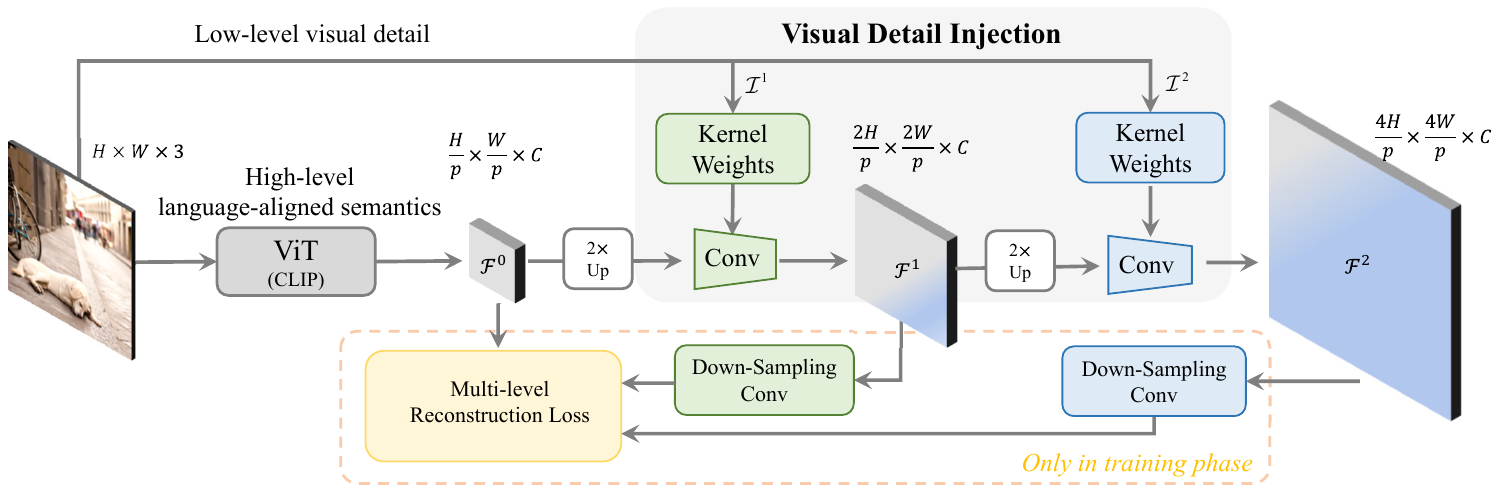}
\caption{
The flowchart illustrates the construction of the Inverse Semantic Pyramid (ISP).
As the first level of ISP, $\mathcal{F}^0$ is the high-level language-aligned semantic features from CLIP-ViT. Subsequent levels, $\mathcal{F}^1$ and $\mathcal{F}^2$, are progressively built by injecting high-frequency visual details from the input image into upsampled features from the previous level, via the Visual Detail Injection Module (VDIM).
A Multi-level Reconstruction (MLR) loss supervises in each scale, ensuring both text-aligned semantic coherence and fine-grained visual fidelity.
}
\vspace{-0.5cm}
\label{fig:featup_module}
\end{figure*}

\subsection{Inverse Semantic Pyramid}\label{subsec:pyramid}
\textbf{Preliminaries.} Traditional convolutional neural networks (CNNs) naturally produce a pyramid of hierarchical bottom-up features $\{\mathcal{F}^l\in\mathbb{R}^{\frac{H}{p\cdot2^l}\times \frac{W}{p\cdot2^l}\times C}\}$. Note that $l$ is the level index, $(H,W)$ is the image resolution, $C$ is the feature dimension, and $p$ is a down-sampling ratio ($e.g.$, $p=8$ in ResNet-50~\cite{he2016resnet}).
In these pyramids, higher-resolution (lower-level) feature maps are rich in visual details and lower-resolution (higher-level) ones contain abstract semantic information. 
However, ViTs, splitting the image into coarse patches, only produce single-scale feature maps ($i.e.$, $\mathcal{F}^0\in\mathbb{R}^{\frac{H}{p}\times \frac{W}{p}\times C}$, where $e.g., p=14$). Lacking such feature pyramids like CNNs hinders their performance on MLLM tasks requiring both fine-grained and high-level visual information. Therefore, how to construct a ViT-based feature pyramid with varying semantic granularity remains a problem.

\smallskip
\noindent\textbf{Visual detail injection module (VDIM).}
Due to the lack of high-resolution features, up-sampling ViT features becomes the necessary strategy to inversely construct the feature pyramid.
Two simple approaches, (1) plain bilinear interpolation and (2) deconvolution network, can be adopted. By doubling and quadrupling the last-layer feature maps, a ViT-based feature pyramid $\{\mathcal{F}^l\in\mathbb{R}^{\frac{H\cdot2^l}{p}\times \frac{W\cdot2^l}{p}\times C}, l=0,1,2\}$ is then constructed. 
While effective, directly up-sampling language-aligned features from CLIP-ViT hardly introduces precise visual details, resulting in suboptimal performance, illustrated in Table.~\ref{tab:mllm_conv_bi_ISP}. To address this, we design a VDIM to up-sample multi-modal semantic features guided by original image priors.

Specifically, the objective of VDIM is to learn $(l-1)$ convolution layers on the image pyramid $\{\mathcal{I}^{l}\in\mathbb{R}^{\frac{H\cdot2^l}{p}\times \frac{W\cdot2^l}{p}\times 3}\}$ to capture high-frequency visual patterns of image texture for guiding the up-sampling process of semantic features, as shown in Fig.~\ref{fig:featup_module}. For each input image, its $(l+1)$-th level features is defined as 

\vspace{-.2in}
\begin{equation}
    \label{eqn:conv}
    \begin{split}
        \mathcal{F}^{l+1}=\mathrm{Conv}\big(\mathrm{Up}(\mathcal{F}^{l}); \Theta^{l+1}(\mathcal{I}^{l+1})\big),
    \end{split}
\end{equation}
where $\mathrm{Up}(\cdot)$ denotes the up-sampling interpolation and $\mathrm{Conv}(\cdot)$ the convolutional operation on feature maps with customized kernel weights $\Theta^{l+1}$ learned on image $\mathcal{I}^{l+1}$.

\smallskip
\noindent\textbf{Optimizing VDIM.}
We propose a multi-level reconstruction (MLR) loss between the higher-level feature maps $\{\mathcal{F}^1,\mathcal{F}^2\}$ and the lowest one $\mathcal{F}^0$ as
\begin{equation}
    \label{eqn:hi_rec_loss}
    \mathcal{L} = \frac{1}{2}\sum_{l=1}^{2}\lVert\mathcal{F}^0-\mathrm{Down}(\mathcal{F}^l;\Omega^{l})\rVert_2^2,
\end{equation}
where $\mathrm{Down}(\cdot)$ is a down-sampling operation with trainable weights $\Omega^l$ in each level. The proposed MLR loss drives the feature pathway to capture low-level textures as well as to maintain multi-modal semantics during the fusion procedure.

\smallskip
\noindent\textbf{Construction of inverse semantic pyramid (ISP).}
As shown in Fig.~\ref{fig:featup_module}, VDIM acts as a progressive feature resolution expanding procedure conditioned on original image priors. 
During the inference, the resulting multi-level feature maps $\{\mathcal{F}^0,\mathcal{F}^1,\mathcal{F}^2\}$ form an ISP, which gathers a hierarchical multi-modal semantic representation with corresponding spatial resolutions supporting proper visual granularity.

\subsection{Hierarchical Window Attention}\label{subsec:window}

The hierarchical nature of ISP necessitates an effective approach for compressing features at varying resolutions while maintaining cross-level spatial alignment. 

\begin{figure}[!t]
\centering
\includegraphics[width=1.0\linewidth]{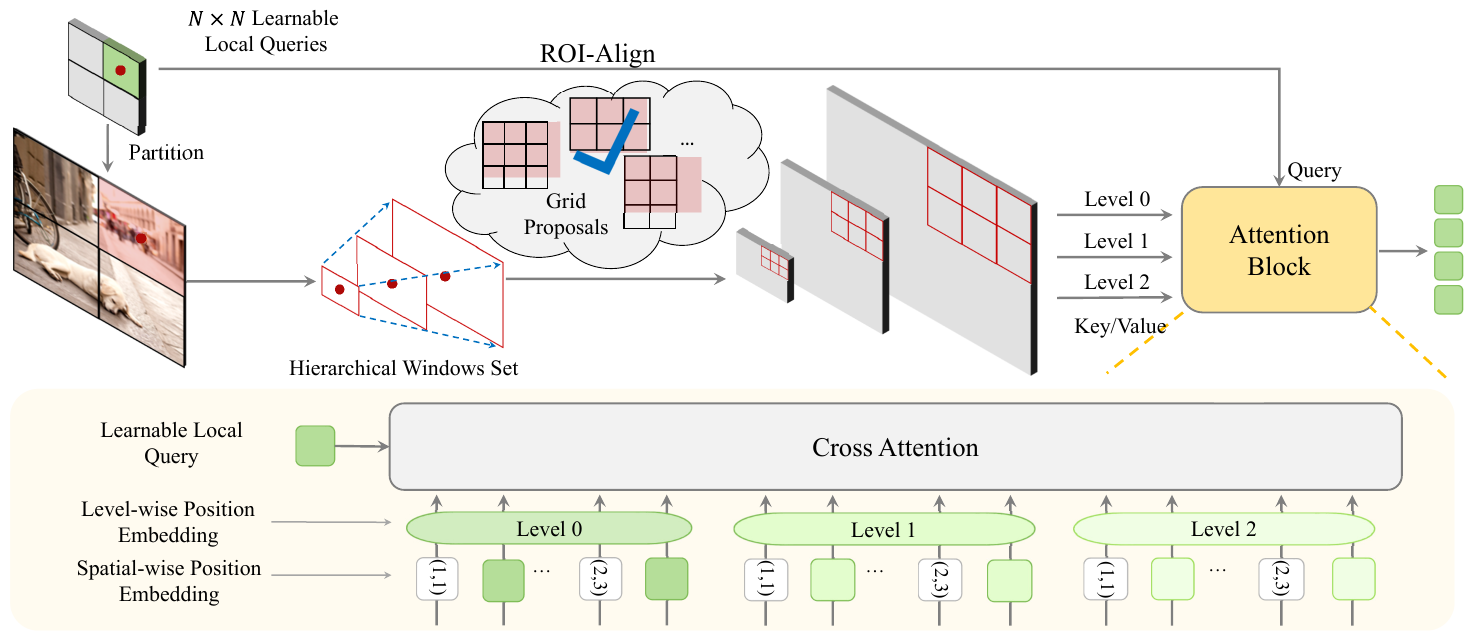}
\caption{The flowchart of hierarchical window attention. 
We initialize a set of learnable queries to attend to local regions. Feature maps from the ISP are processed by a set of cross-scale windows, forming hierarchical and local-aware features at different levels. The features are then concatenated along the length axis, to serve as the key and value for the learnable queries. The output is condensed visual tokens rich in diverse and local-aware semantics.}

\vspace{-0.5cm}
\label{fig:hiwin-attn}
\end{figure}

\smallskip
\noindent\textbf{Hierarchical window generation.}\label{win-gen}
Inspired by object detection~\cite{ren2015faster,zhu2020deformable}, we utilize the RoI-align~\cite{he2017mask} to sample key features to keep the spatial locality of cross-level feature maps, in Fig.~\ref{fig:hiwin-attn}. Specifically, we first uniformly divide feature maps of each level into $N\times N$ windows, whose widths and heights are float-point values $(\frac{W}{N},\frac{H}{N})$ rather than integers.
Windows share the same ``anchor" point form a set of hierarchical bounding boxes (coordinates of top-left and bottom-right) $\{\mathcal{R}^l_{i,j}\in \mathbb{R}^{1\times4}, i,j\in0,1,2...N-1,l=0,1,2\}$, where $l$ is the feature level and $(i,j)$ the 2D index. 
To mitigate the size distortion of feature maps caused by a difference between the aspect ratio of RoI-aligned feature maps and the image, we define a pooling score to evaluate this difference: 
\begin{equation}
    \label{eq:regionslice}
    \begin{split}
        S(W,H,r_{w},r_{h})= -\left| \log \frac{W}{H} - \log \frac{r_{w}}{r_{h}}\right|,
    \end{split}
\end{equation}
where $(r_{w},r_{h})$ denotes the width and height of pooled features.
By maximizing the score $S$, we select the optimal grid size $(r^*_{w},r^*_{h})$ from pre-defined proposals $\{(3,3),(2,3),(3,2),(2,4),(4,2)\}$. 
Then, we carry out the RoI-align with the generated windows to sample the key feature maps for the following attention operation.

\smallskip
\noindent\textbf{Cross-scale window querying.}
To compress the ISP $\{\mathcal{F}^l,l=0,1,2\}$ of one image or slice $\mathcal{I}$, we initialize a set of queries $\{\mathcal{Q}_{i,j}\in \mathbb{R}^{1\times C}, i,j\in0,1,2...N-1\}$, each of which corresponds to a set of hierarchical windows $\{\mathcal{R}^l_{i,j}, l=0,1,2\}$, in Fig.~\ref{fig:hiwin-attn}. Regarding each query vector $\mathcal{Q}_{i,j}$, we prepare the key vector $\mathcal{K}^{l}_{i,j}\in\mathbb{R}^{(r^*_{w}\cdot r^*_{h})\times C}$ of $l$-th level as
\begin{equation}
\begin{split}
    \mathcal{K}^{l}_{i,j} = \text{RoI}(\mathcal{F}^l, \mathcal{R}^l_{i,j}) + \phi^{l},
\end{split}
\end{equation}
where $\phi^{l}$ is a level positional embedding. We then concatenate the $\mathcal{K}^{l}_{i,j}$ in length axis and form the final key vector $\mathcal{K}_{i,j}\in \mathbb{R}^{(3\cdot r^*_{w}\cdot r^*_{h})\times C}$ for each $\mathcal{Q}_{i,j}$. The corresponding value vector $\mathcal{V}_{i,j}$ is obtained in the same way yet without the level positional embedding. Thus, the cross-attention can be performed as
\begin{equation}
    \mathcal{Q^*}_{i,j} = \text{CrossAttn}(\mathcal{Q}_{i,j}+\varphi_{i,j}, \mathcal{K}_{i,j}+\zeta_{i,j}, \mathcal{V}_{i,j}),
\end{equation}
where $\mathcal{Q^*}_{i,j}$ denotes the updated query, and $\varphi,\zeta$ is the 2D spatial position embedding of query and key vector, respectively. In the end, we concatenate all the $\mathcal{Q^*}_{i,j}$ into a feature map $\mathcal{P}\in \mathbb{R}^{N\times N\times C}$ to represent the visual token of $\mathcal{I}$.

\subsection{Spatially-consistent Token Organization}

Due to the varying partitions of different images, organizing and conveying the structure of slices to the LLM facilitates a more accurate understanding of the image. Prior studies~\cite{guo2024llava-uhd,SPHINX2023,yao2024minicpm} utilize spatial tokens (e.g., “$\backslash n$” or “,”) to denote the relative positioning of image slices. While such approaches provide cues regarding the global arrangement of each slice, they overlook the intrinsic spatial relationships among individual visual tokens. For instance, two horizontally adjacent tokens in the image, located in different slices, may become significantly separated in a 1D arrangement. Leveraging our Hiwin transformer, which preserves the 2D spatial consistency with the original image, we amalgamate these 2D feature maps into a large 2D feature map according to the slicing configuration, in Fig.~\ref{fig:pipe}. 
\section{Experiment} 
\label{sec:experiment}
In this section, we conduct an empirical evaluation of LLaVA-UHD v2. We begin with a comprehensive outline of the implementation details of our model, followed by a comparative analysis of its performance across widely recognized benchmarks against competitive counterparts. Finally, we provide an in-depth ablation to further elucidate the capabilities and behaviors of LLaVA-UHD v2.

\subsection{Implementation Details}
\textbf{Model Setting.}
We adopt LLaVA-UHD~\cite{guo2024llava-uhd} as the baseline method. Specifically, we employ CLIP-ViT-L/14-336 as the visual encoder, Vicuna-7B/13B~\cite{chiang2023vicuna} or Qwen2-7B~\cite{wang2024qwen2} as the language model, 
and our proposed Hiwin transformer as the vision-language projector. 
We set the maximum slice number to 6 to cover a range of aspect ratios and image resolutions. The number of learnable local queries is set as 144 ($i.e.,$$N=12$). 
Before employing the VDIM in MLLM, we pre-train it with frozen CLIP-ViT on MS-COCO~\cite{lin2014mscoco} with a global batch of 16 on 8$\times$A100. We leverage Adam optimizer with 1$e^{-3}$ learning rate for 2000 steps. 
This is an independent phase for building a task-agnostic representation, and the weights of the ISP are always reused. 
LLaVA-UHD v2 consists of a two-stage multi-modal training process as outlined below. 

\smallskip
\noindent\textbf{Stage 1: MLLM pre-training.} 
In this stage, the parameters of the visual encoder, pre-trained VDIM, and LLM are frozen. We only fine-tune the parameters within the hierarchical window attention of the HiWin transformer using LLaVA-Pretrain~\cite{liu2023llava1.5} for 1 epoch with a global batch size of 256. We employ the AdamW optimizer and a cosine learning rate scheduler. The learning rate is 1e$^{-3}$ for Vicuna-7B, 2e$^{-4}$ for Vicuna-13B, and Qwen2-7B. Note that, in this stage, we only encode the overview image without the slices for efficiency.

\smallskip
\noindent\textbf{Stage 2: MLLM supervised fine-tuning.} 
In this stage, we fine-tune all parameters except those in the VDIM. The learning rate is set to 2e$-5$ with a batch size of 128. To manage training costs, we use 825k data for analysis and ablation studies, including LLaVA-mix665k~\cite{liu2023llava1.5} and 160k from Ureader~\cite{ye2023ureader}. 
For comparison with advanced MLLMs, we balance our data distribution and introduce an 858k-mixed dataset, which was detailed in supplementary. 

\subsection{Experimental Setting}
We present the experimental settings, detailing the benchmarks, evaluation metrics, and compared counterparts.

\smallskip
\noindent\textbf{Benchmarks.} 
Extensive benchmarks are used to analyze the effect of our modules. We categorize these benchmarks into the following folds: (1) General VQA benchmarks including MME~\cite{fu2023mme}, MMB~\cite{liu2023mmbench}, SEED-Image~\cite{li2023seed}, GQA~\cite{hudson2019gqa}, MMStar~\cite{chen2024we} and HallusionBench:~\cite{guan2024hallusionbench} ; (2) Knowledge-based VQA benchmarks including MMMU-val~\cite{yue2024mmmu}, Science-QA~\cite{lu2022scienceqa}, AI2D~\cite{kembhavi2016AI2D}, MathVista~\cite{lu2023mathvista}; (3) OCR-based VQA benchmarks including ChartQA~\cite{masry2022chartqa}, OCR-Bench~\cite{liu2023ocrbench}, TextVQA~\cite{singh2019textqa} and DocVQA~\cite{DocVQA2021};  (4) Visual spatial understanding benchmarks such as RealWorldQA~\cite{realwordqa} and RefCOCOs~\cite{yu2016rec}; (5) High-resolution image perception benchmarks like HR-Bench(4K)~\cite{wang2024hrbench}.

\smallskip
\noindent\textbf{Evaluation Protocols.}
Beyond benchmark evaluations, we report additional metrics for comprehensive analysis: (1) overall volume of training data, (2) maximum supported image resolution for each method, and (3) computation cost of the entire MLLM at maximum resolution.

\noindent\textbf{Counterparts.} 
We compare our model with the advanced MLLM counterparts. (1) General MLLMs like Honey-bee~\cite{cha2024honeybee}, Dense Connector~\cite{yao2025dense}, VILA~\cite{lin2024vila} and LLaVA-1.5~\cite{liu2023llava1.5}. (2) High-resolution MLLMs including Monkey~\cite{li2023monkey}, LLaVA-Next~\cite{2023llava1.6}, PIIP-LLaVA~\cite{wang2025parameter}, SliME-Llama3-8B~\cite{zhang2024beyond}, DeepseekVL-7B~\cite{lu2024deepseek} and Token-Packer~\cite{li2024tokenpacker}. (3) Mixture of visual experts such as LLaVA-HR~\cite{luo2024llavahr}, SPHINX-series~\cite{SPHINX2023,gao2024sphinxx} , MG-LLaVA~\cite{zhao2024mg} and Mini-Gemini~\cite{anil2023gemini}. (4) OCR-centric MLLMs including UReader~\cite{ye2023ureader} and TextMonkey~\cite{liu2024textmonkey}.

\begin{table*}[tb]
  \setlength{\tabcolsep}{1.0pt}
  \caption{Main performance on popular benchmarks. 
  \#Data denotes the volume of overall data during MLLM pre-training and supervised fine-tuning. ``MaxRes." is the maximum accessible resolution of MLLM. ``VQA$^\mathrm{D}$": DocVQA. ``Bench$^{\mathrm{OCR}}$": OCR-Bench. ``VQA$^{\mathrm{C}}$": ChartQA. ``VQA$^\mathrm{T}$": TextVQA. ``SQA": Science-QA. ``MMMU$^\mathrm{v}$": MMMU-val. ``Math.": MathVista. ``SEED$^\mathrm{I}$": SEED-Image. ``MME$^\mathrm{P}$": perception sub-set of MME. ``RWQA": RealWorldQA. ``Bench$^{\mathrm{HR}}$": HR-Bench. }
  \centering
  \fontsize{7.2}{8.5}\selectfont
  {
  \scalebox{0.70}{
  \begin{tabular}{@{}ccccc|cccc|cccc|ccccc|c|ccccccccccc@{}}
    \toprule
    & & &  &  &\multicolumn{4}{c|}{OCR \& Chart} & \multicolumn{4}{c|}{Knowledge} & \multicolumn{5}{c|}{General}  &\multicolumn{1}{c|}{\makecell{Vision \\ Spatial}}  &\makecell{High Res.} \\
    \multirow{1}{*}{Method} & \multirow{1}{*}{LLM} &\multirow{1}{*}{\#Data} &\multirow{1}{*}{MaxRes.} &\multirow{1}{*}{\#FLOPs.}&VQA$^\mathrm{D}$&Bench$^{\mathrm{OCR}}$&VQA$^\mathrm{C}$ &VQA$^\mathrm{T}$ &AI2D &SQA&MMMU$^\mathrm{v}$ &Math.&GQA &SEED$^\mathrm{I}$&MMB &MME$^\mathrm{P}$&MMStar&RWQA &Bench$^{\mathrm{HR}}$ \\
    \toprule
    mPLUG-Owl2~\cite{ye2023mplugowl2revolutionizingmultimodallarge}& Llama2-7B &401M &448$\times$448 & 1.7T&-&-&-&58.2&-&68.7&-&25.5&56.1&57.8&64.5&72.5&34.8&-&-\\
    UReader~\cite{ye2023ureader} & Llama2-7B&86M&896$\times$1120  &20.3T&65.4&-&59.3&57.6&-&-&-&-&-&-&-&-&- &-&-   \\
    VILA~\cite{lin2024vila} & Llama2-7B&51M&336$\times$336  &8.2T &-&-&-&64.4&-&68.2&-&-&62.3&61.1&68.9&76.7&-&-&- \\ 
    SPHINX-2k~\cite{SPHINX2023} & Llama2-Accessory&1.01B  &762$\times$762  &42.2T&-&-&-&61.2&-&70.6&-&-&63.1&71.6&65.9&73.6&-&-&-\\
    SPHINX-X~\cite{gao2024sphinxx} & Llama2-Accessory&15.3M&448$\times$448&21.3T&56.3&-&39.7&58.1&63.0&70.4&-&-&56.2&68.8&57.9&63.0&-&-&-\\
    LLaVA-HR~\cite{luo2024llavahr} & Vicuna-7B&1.22M&1024$\times$1024&24.3T&-&-&-&67.1&-&65.1&-&-&64.2&64.2&-&77.7&-&-&-\\
    Honey-bee~\cite{cha2024honeybee} & Vicuna-7B&52.5M&336$\times$336&2.6T&-&-&-&-&-&-&35.3&-&-&64.5&70.1&77.2&-&-&-\\
    Mini-Gemini~\cite{li2024minigemini} & Vicuna-7B&3.0M&672$\times$672&54.6T&61.9&47.7&47.4&65.2&68.2&69.6&36.8&-&64.5&66.9&65.8&77.3&-&51.1&50.1\\
    Monkey~\cite{li2023monkey} & Vicuna-7B&1.40B  &896$\times$1344  & 28.0T&66.5&51.4&65.1&67.6&62.6&69.4&38.9&33.5&60.7&64.3&59.8&73.6&37&51.6&38.0    \\
    LLaVA-1.5~\cite{liu2023llava1.5} & Vicuna-7B&1.22M  & 336$\times$336  & 8.0T&21.8&31.8&17.8&45.5&55.5&66.8&37.0&25.5&62.0&65.8&66.5&75.3&33.1&54.8&36.1  \\
    LLaVA-Next~\cite{2023llava1.6} & Vicuna-7B&1.34M&672$\times$672&44.4T&63.6&53.2&54.3&64.9&67.0&70.1&35.8&34.6&64.2&70.2&67.4&76.0&37.6&57.8&47.9 \\
    Token-Packer~\cite{li2024tokenpacker} & Vicuna-7B&2.7M&1008$\times$1008&13.1T&60.2&45.2&-&68.0&-&-&35.4&-&-&67.4&74.5&-&-&- \\
    TextMonkey~\cite{liu2024textmonkey} & Vicuna-7B&1.45B  &448$\times$448 & 4.0T&66.7&-&59.9&64.3&-&-&-&-&-&-&-&-&- &-  &-  \\
    LLaVA-1.5~\cite{liu2023llava1.5} & Vicuna-13B&1.22M  & 336$\times$336  & 15.1T&-&-&-&61.3&-&71.6&-&-&63.3&61.6&67.7&76.5&-&-&-  \\
    LLaVA-Next~\cite{2023llava1.6} & Vicuna-13B&1.34M&672$\times$672&67.0T&-&53.7&61.4&67.1&-&73.6&36.2&35.3&65.4&71.9&70.0&76.5&40.4&57.6&- \\
    LLaVA-1.5-Qwen~\cite{an2024multi} & Qwen2-7B&1.22M  & 336$\times$336  & 8.2T&-&-&-&-&64.9&-&40.7&33.6&62.7&69.4&72.0&76.0&-&-&-  \\
    Dense Connector~\cite{yao2025dense} & Llama3-8B&1.22M&384$\times$384&11.6T&-&-&-&-&-&75.2&40.4&28.6&65.1&-&74.4&-&-&-&- \\ 
    LLaVA-LLaMA3~\cite{2023xtuner} & Llama3-8B& 1.22M & 336$\times$336 &8.7T&-&-&-&-&-& 73.3 & 36.8 &-& 63.5&-& 68.9 & - &- &- &-  \\ 

    PIIP-LLaVA~\cite{wang2025parameter} & Llama3-8B& 1.22M & 1024$\times$1024 &36.0T&-&-&-&67.1&-& 68.3 & - &-& 63.9&69.4& 67.0 & - &- &- &-  \\ 
    MG-LLaVA~\cite{zhao2024mg} & Llama3-8B& 2.5M & 768$\times$768 &33.0T&-&-&-&67.3&-& 70.8 & - &-& -&69.4& 72.1 & - &- &- &- \\ 
    SliME~\cite{zhang2024beyond} & Llama3-8B& 2.0M & 2016$\times$2016 &62.0T&-&-&-&64.8&-& - & 41.2 &-& 63.9& - & 75.0 & - &- &- &- \\ 
    DeepseekVL-7B~\cite{lu2024deepseek} & DeepseekLLM-7B& - & 1024$\times$1024 &-&-&45.6&-&-&-& - & 36.6 &36.1& -& 70.4 & 73.2 & - &37.1 & - &-   \\ 
    \midrule\midrule
    LLaVA-UHD v2 &Vicuna-7B&1.42M&1008$\times$672&17.5T&68.1 &53.9 &64.5 &67.6 &70.5 &71.3 &38.2 &34&65.4 &70.0 &68.2 & 74.7 & 40.2 &58.2 &51.5\\
    LLaVA-UHD v2 &Vicuna-13B&1.42M&1008$\times$672&26.4T&68.2 &55.6 &67.4 &70.0 &72.4 &73.3 &37.7 &35.2&\textbf{66.0} &71.1 &70.3 & 73.1 & 42.0 &59.6 &55.3\\
    LLaVA-UHD v2 & Qwen2-7B&1.42M&1008$\times$672 & 14.3T & \textbf{72.9} & \textbf{57.7} & \textbf{70.4} & \textbf{70.6} & \textbf{75.5} & \textbf{76.9} & \textbf{43.3} &\textbf{39.1}& 65.1 & \textbf{73.6} & \textbf{77.1}  & \textbf{78.8} & \textbf{49.4} & \textbf{64.6} & \textbf{59.9} \\

  \bottomrule  
  \vspace{-0.5cm}
  \end{tabular}
  }
  }
  
\label{tab:sota}
\end{table*}

\begin{table*}[tb]
  \setlength{\tabcolsep}{2.0pt}
  \caption{Ablation studies of modules in our proposed method.  ``$\Delta$" denotes the overall improvement compared to the baseline. REC reports the average accuracy of RefCOCO/g/+.}
  \centering
  \fontsize{7.8}{9.0}\selectfont
  \scalebox{0.82}{
  {
  \begin{tabular}{c|c|cccc|ccc|cccc|cc|ccccccccc}
    \toprule
     &  & \multicolumn{4}{c|}{OCR \& Chart}  & \multicolumn{3}{c|}{Knowledge} & \multicolumn{4}{c|}{General}
      &\multicolumn{2}{c|}{\makecell{Vision Spatial}}&\makecell{High Res.} \\
    \multirow{1}{*}{Method}&Average &VQA$^{\mathrm{D}}$&Bench$^{\mathrm{OCR}}$&VQA$^{\mathrm{C}}$ &VQA$^{\mathrm{T}}$&AI2D&SQA&MMMU$^{\mathrm{v}}$&GQA&SEED$^{\mathrm{I}}$&MMB&MME$^{\mathrm{P}}$&RWQA&REC&Bench$^{\mathrm{HR}}$ \\
    \toprule
    \makecell[c]{LLaVA-UHD~\cite{guo2024llava-uhd}} &58.0 &56.7 &40.9 &56.3 &62.2 &55.4 &70.7 &37.0 &63.8 &65.6 &64.8 &70.0 &54.4 &68.3  &45.6\\
    \makecell[l]{$+$ \textit{VDIM(ISP)}}&60.0&60.2 &50.4 &60.4 &67.1 &57.8 &70.5 &38.2 &64.0 &66.7 &65.6 &71.2 &51.9 &72.3 &43.9 \\
    \makecell[l]{$+$ \textit{Hiwin attention}} &61.5&65.0 &51.3 &62.5 &68.5 &58.1 &69.2 &38.9 &64.6 &67.4 &65.5 &73.0 &55.5 &73.3  &48.9 \\
    \makecell[l]{$+$ \textit{Token organization}} &61.7&66.0 &50.1 &62.8 &66.8 &59.4 &69.8 &37.6 &64.0 &67.4 &66.1 &73.6 &56.9 &74.0 &49.0\\
    \midrule
    $\Delta$ &+3.7 &+9.3 &+9.2 &+6.5 &+4.6 &+4.0 &-0.9 &+0.6 &+0.2 &+1.8 &+1.3 &+3.6 &+2.5 &+5.7 &+3.4\\
    \toprule
  \end{tabular}
  }
  }
\vspace{-0.5cm}  
\label{tab:module-ablation}
\end{table*}

\subsection{Main Performance}
Table~\ref{tab:sota} showcases a comparative analysis of our proposed LLaVA-UHD v2 against state-of-the-art MLLMs across 15 widely recognized benchmarks. 
\textbf{(1) LLaVA-UHD v2 outperforms current counterparts}. 
Compared with general models (such as LLaVA-1.5, Dense Connector) and high-resolution MLLMs (like PIIP-LLaVA, SliME-Llama3-8 and DeepseekVL-7B), LLaVA-UHD v2 demonstrates consistent improvements across various tasks, including general VQA (e.g., 77.1\% on MMB and 49.4\% on MMStar), ultra-high-resolution image perception (e.g., 59.9\% on HR-Bench). Notably, LLaVA-UHD v2 surpasses OCR-centric models like TextMonkey on DocVQA (72.9\% vs. 66.5\%) and outperforms those with multiple experts (such as MG-LLaVA), achieving superior performance on general tasks like SEED (73.6\%). These results underscore the value of rich semantics derived from multi-level multi-modal granularity, enhancing both the understanding and perception abilities of MLLMs.
\textbf{(2) LLaVA-UHD v2 indicates efficiency on data utilization and computation}. Compared to LLaVA-Next and Mini-Gemini, both operating at a 672$\times$672 resolution, LLaVA-UHD v2 supports 1.5 times the resolution ($i.e$., 672$\times$1008) and achieves superior performance with less than 40\% of the computational cost. Furthermore, in contrast to Honey-bee and VILA, which utilize 52.5M and 51M data samples respectively, LLaVA-UHD v2 attains comparable or superior performance using only $\sim$2.8\% of the data, demonstrating the data efficiency of our model. As for the training duration, under the same model configuration and data volume, LLaVA-UHD v2 requires $\sim$27 hours to train on 8$\times$A100 GPUs, while LLaVA-Next needs $\sim$42 hours, which is well-suited for low-cost exploratory research in the academic community.


\subsection{Analytical Study}
We conduct analytical experiments on the proposed modules to verify the effect of LLaVA-UHD v2. Without special instructions, we use Vicuna-7B as the base LLM.

\smallskip
\noindent\textbf{Main module ablation.}
In Table~\ref{tab:module-ablation}, by replacing the low-resolution CLIP-ViT features with highest-level ones of the inverse semantic pyramid (ISP) constructed by VDIM, 2.2\% average improvement can be seen, especially, on tasks depending on visual details like OCR-Bench (+9.5\%) and RefCOCOs (+4.0\%). Employing the Hiwin attention to integrate the ISP further increases 1.3\% in average accuracy, especially 5.0\% on ultra high-resolution perception (HR-Bench), demonstrating rich visual granularity could facilitate precise language generation. With spatially consistent token organization, visual-spatial understanding observes further improvement, such as 1.3\% and 0.6\% on RealWorldQA and RefCOCOs, respectively.




\begin{table}[!tb]
  \setlength{\tabcolsep}{2.0pt}
  \caption{Comparison of different methods for semantic pyramid construction. ``\textit{ConvNext}" means we replace the CILP-ViT with CLIP-ConvNext~\cite{liu2022convnet} as visual encoder and directly use the feature maps from multiple stages as the final hierarchical feature pyramid.
  }
  \centering
  \footnotesize
  {
  \begin{tabular}{c|c|cc|c|ccc|c}
    \toprule
     & & \multicolumn{2}{c|}{General} & \multicolumn{1}{c|}{\makecell{Knowledge}} & \multicolumn{3}{c|}{\makecell{OCR \& Chart}}  &\makecell{High Res.} \\
    \multirow{1}{*}{Method} &Average &MME$^{\mathrm{P}}$ &GQA &AI2D &VQA$^{\mathrm{C}}$ &VQA$^{\mathrm{T}}$ &VQA$^{\mathrm{D}}$ &Bench$^{\mathrm{HR}}$ \\
    \toprule
    LLaVA-UHD  & 59.9 &70.0 &63.8 &55.4 &56.3 &62.2 &56.7 &45.6 \\
    \makecell[l]{w/  \textit{ConvNext}} &59.7 &68.2 &62.7 &55.6 &61.8 &63.5 &61.8 &44.0 \\
    \makecell[l]{w/  \textit{DeConv.}} &61.7 &71.2 &64.2 &57.4 &61.8 &67.8 &63.4  &46.3 \\
    \makecell[l]{w/  \textit{Bilinear}} &62.0 &72.0 &64.5 &57.8 &62.2 &67.6 &63.7 &46.5 \\
    \makecell[l]{w/  \textit{VDIM}} &63.0 &73.0 &64.6 &58.3 &62.5 &68.5 &65.0 &48.9\\
    \toprule
  \end{tabular}
  }
\vspace{-0.5cm}
\label{tab:mllm_conv_bi_ISP}
\end{table}

\smallskip
\noindent\textbf{ISP demonstrates effectiveness on MLLM tasks compared to traditional feature pyramid.}
In Table.~\ref{tab:mllm_conv_bi_ISP}, it is evident that feature pyramids, regardless of their construction method, could enhance performance across various MLLM tasks. Nonetheless, the ISP constructed by VDIM achieves an average performance gain of 1.0\% over bilinear interpolation, indicating that the ISP further enhances beneficial visual representations ($e.g.$, high-frequency visual details).

\vspace{-4.5mm}
\begin{figure}[h] 
    \centering
    \begin{subfigure}[b]{0.60\textwidth}
        \centering
        \includegraphics[width=\textwidth]{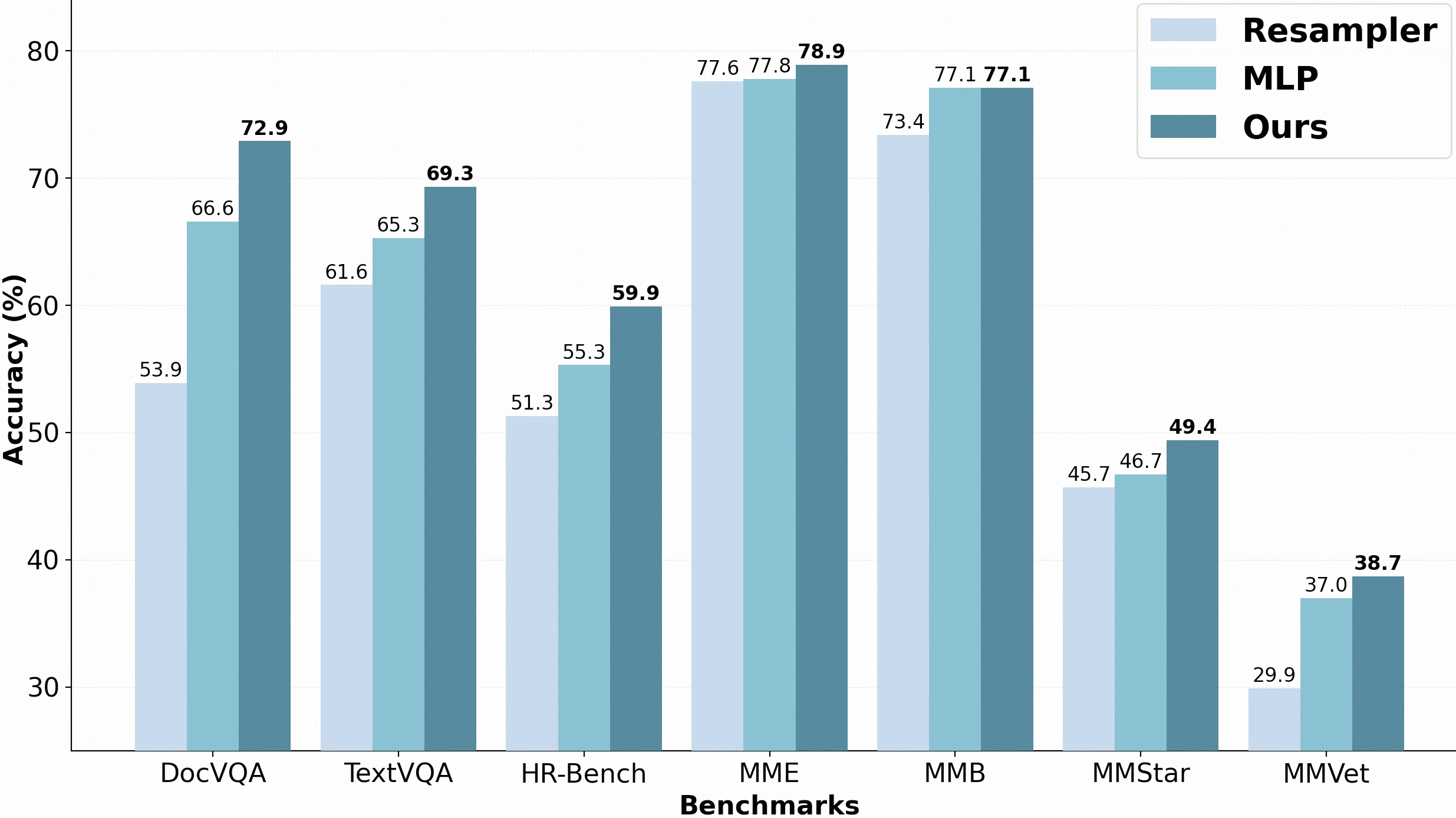} 
        \caption{}
        \label{fig:ablation_projector}
    \end{subfigure}
    \hfill
    \begin{subfigure}[b]{0.35\textwidth}
        \centering
        \includegraphics[width=\textwidth]{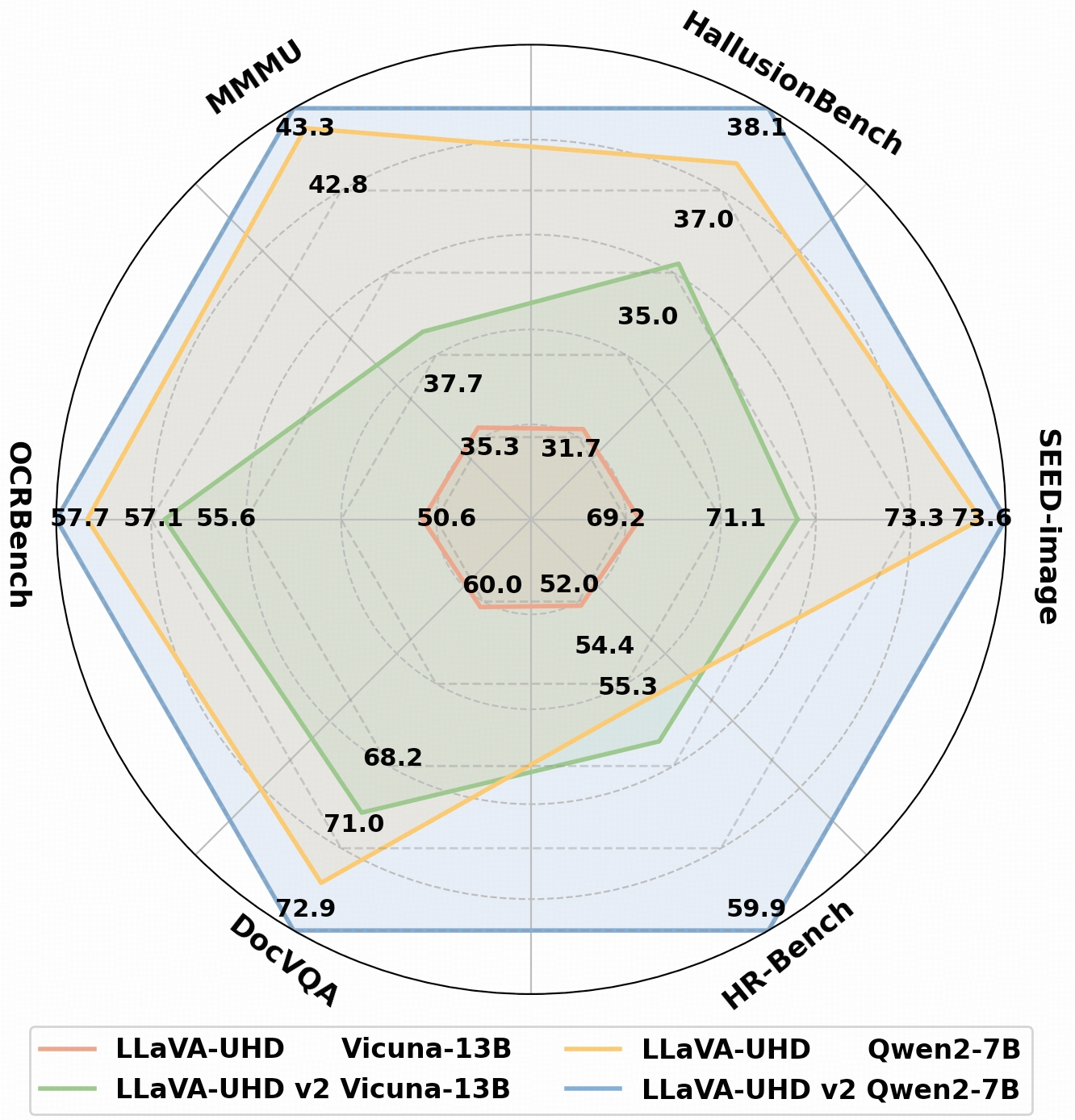} 
        \caption{}
        \label{fig:mllm_ablation_radar}
    \end{subfigure}
    \caption{\textbf{Comparison of performance}. (a) Performance of using different projectors on compressing ISP. Hiwin attention exhibits a significant advantage. (b) Performance of our model equipped with different LLMs. }
    \label{fig:main}
\end{figure}

\smallskip
\noindent\textbf{Hierarchical window attention works on compressing multi-scale feature pyramid}.
As shown in Fig.~\ref{fig:ablation_projector}, Hiwin attention condenses visual tokens by using learnable queries on feature pyramids with arbitrary native resolutions, resulting in superior performance on fine-grained MLLM tasks (e.g., DocVQA and TextVQA).
In contrast, directly down-sampling high-resolution features via interpolation before MLP projection~\cite{liu2023llava1.5} loses critical visual details and thereby degrades performance. Moreover, the Perceive Resampler~\cite{Alayrac2023Flamingo,yao2024minicpm} performs notably worse due to its lack of spatial prior constraints on each query, which adversely affects training convergence, as evidenced by its significantly higher pre-training loss compared to Hiwin attention (0.6110 vs. 0.4368).

\noindent\textbf{Hiwin transformer keeps superior when transferring to other LLMs}. 
In Fig.~\ref{fig:mllm_ablation_radar}, the proposed Hiwin transformer consistently improves performance across Vicuna-13B, and Qwen2-7B, demonstrating its strong generalization capability to different model scales and architectures. Especially in tasks that require visual details, such as DocVQA and HR-Bench, the improvement is even more pronounced.

\subsection{Visualization Analysis}

\begin{figure*}[!t]
\centering
\includegraphics[width=1.0\linewidth]{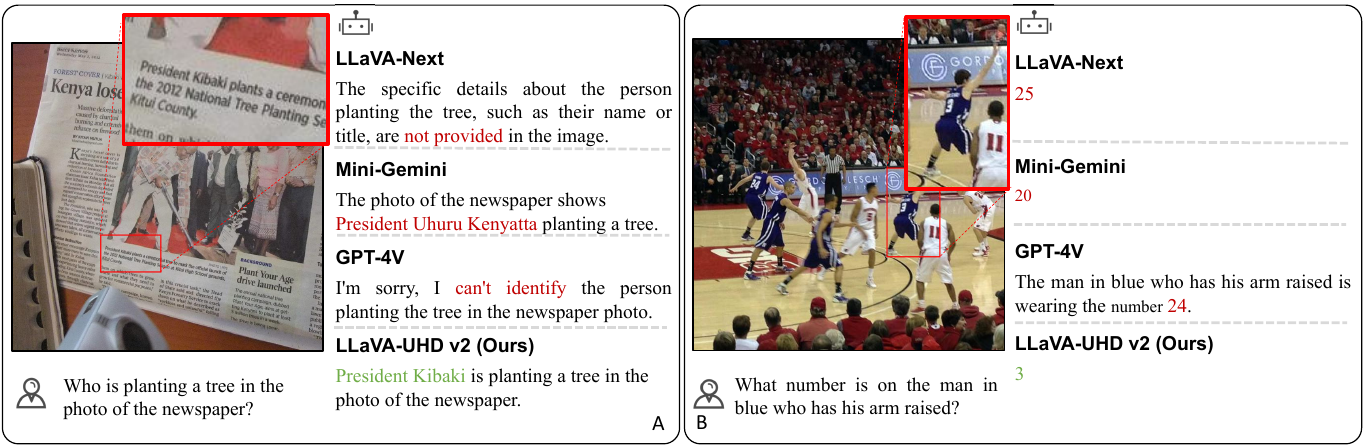}
\caption{
Qualitative comparison of proposed LLaVA-UHD v2 and advanced MLLMs, including LLaVA-Next, Mini-Gemini, and GPT-4V. Our method outperforms its counterparts, by providing both fine-grained visual information and high-level semantic contexts for the high-resolution complex perception tasks
}
\vspace{-0.3cm}
\label{fig:case_study}
\end{figure*}

\begin{figure}[!t]
\centering
\includegraphics[width=1.0\linewidth]{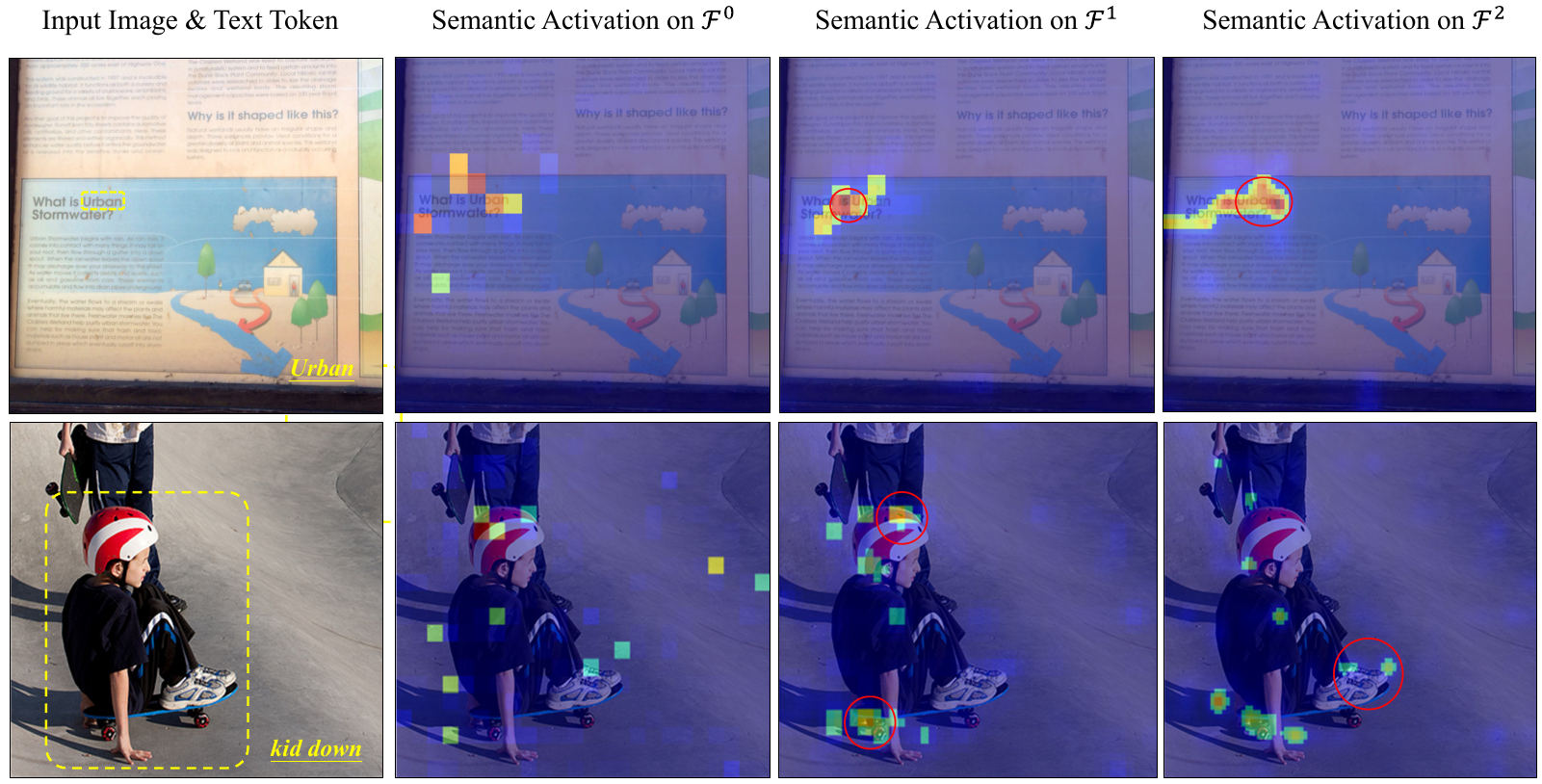}
\caption{Activation response of specific textual tokens to different visual feature levels, exhibiting complement to each other. Red circles highlight the obvious difference between levels (Best viewed in color and zoomed-in).
}
\vspace{-0.5cm}
\label{fig:semantic_activation}
\end{figure}

\textbf{Case study.}
{In Fig.~\ref{fig:case_study}, we visualize the performance of well-known MLLMs on high-resolution, complex perception tasks. This kind of task requires MLLMs to well fuse both visual details and high-level semantics to accurately identify fine-grained targets ($e.g.$, OCR, colors) during the procedure of complex semantic perception ($e.g.$, semantic relation and visual behavior). It is evident that LLaVA-UHD v2 correctly recognizes the tree planter in the newspaper photo and associates it with the name within the dense image caption (Case A). We also can see that LLaVA-UHD v2 captures the player who raises his hands and reads the ``number 3'' on his clothes (Case B).
In contrast, LLaVA-Next overlooks the name information within dense texts (Case A) and hallucinates on the player number (Case B). Mini-Gemini fails to extract the true name (Case A) and also hallucinates (Case B). Additionally, GPT-4V shows limitations in referencing the information in the newspaper (Case A) and falsely recognizes ``number 24'' due to wrong fine-grained action perception (Case B).}

\smallskip

\noindent\textbf{Semantic activation cross semantic scale.}
In Fig.~\ref{fig:semantic_activation}, we demonstrate the activation responses of specific textual prompts in the language model to the inverse semantic pyramid. As shown, OCR-like textual tokens yield finer-grained and more accurate activations at higher feature levels, facilitating accurate scene text recognition (first row). For object-level semantics, higher feature levels enhance edge detail activations, enabling more precise semantic localization (second row). Collectively, the semantic pyramid offers a more exhaustive set of visual semantics with rich granularity, effectively supporting nuanced language decoding.

\section{Conclusion}
\label{sec:conclusion}

Our proposed hierarchical window transformer, which is the core of LLaVA-UHD v2, effectively addresses the limitations of conventional ViT-based MLLMs by capturing varying visual granularity essential for precise language generation. 
The Hiwin transformer adeptly constructs an inverse semantic pyramid for enriched multi-modal representation, which is then condensed into a compact set of visual tokens. This process enhances nuanced visual-linguistic alignment as well as facilitates efficient visual prompting for the LLM. LLaVA-UHD v2 shows substantial gains over the baseline method across a range of MLLM benchmarks, demonstrating its capacity in MLLM tasks that demand both low-level details and high-level semantics. Furthermore, the Hiwin transformer offers versatility, presenting potential adaptability across diverse ViT-based MLLM architectures.

{
    \small
    \bibliographystyle{nat_fullname}
    \bibliography{main}
    
}

\newpage

%

\clearpage
\setcounter{section}{0}
\section{Appendix}
\appendix

In the supplemental materials, we report the experimental performance of the proposed inverse semantic pyramid (ISP) on general visual tasks, the details of the visual detail injection module (VDIM), and the dataset composition in the supervised fine-tuning phase of MLLM. 
Furthermore, we additionally analyze the behaviors of LLaVA-UHD v2 through qualitative and quantitative experiments.

\section{Implementation Details}

\subsection{Experimental setting on visual tasks}

\begin{figure}[!t]
\centering
\scalebox{0.74}{
\includegraphics[width=1.0\linewidth]{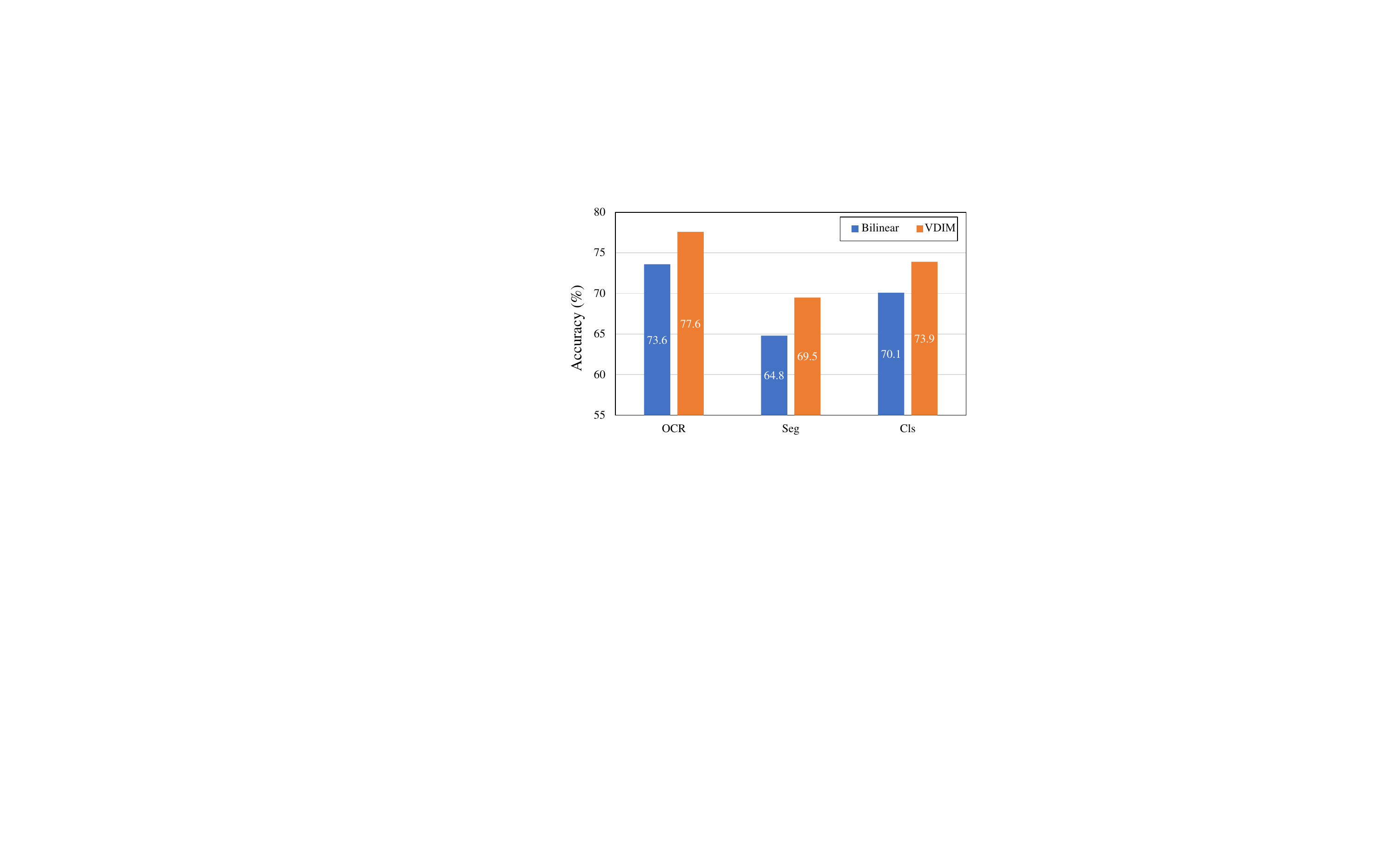}}
\caption{Performance on different visual tasks with VDIM and vanilla bilinear interpolation. ``OCR" denotes the optical character recognition, ``Seg" the Linear probing semantic segmentation, and ``Cls" the fine-grained classification on SUB-200.
}
\label{fig:cv_jbu_bi_deconv}
\end{figure}

\textbf{Results on visual tasks.} Beyond the VQA task in MLLM, we further evaluate some fundamental visual tasks, including semantic segmentation~\cite{lin2014mscoco}, optical character recognition~\cite{mishra2012iit5k}, and fine-grained classification~\cite{keskar2016cub}, to compare the effect of bilinear interpolation and VDIM.
In Fig.~\ref{fig:cv_jbu_bi_deconv}, the VDIM outperforms the bilinear interpolation on OCR (+4.0\%), semantic segmentation (+4.7\%) and fine-grained classification (+3.8\%), which demonstrate more visual details are encoded for precise semantic discrimination. The implementation details for each visual task are provided below.
Unless explicitly stated, both naive bilinear interpolation and the VDIM are employed to perform feature up-sampling on low-resolution (\textit{i.e.}, 24$\times$24) feature maps which are extracted from the second-to-last layer of CLIP-ViT~\cite{radford2021clip}.

\smallskip
\textbf{Optical character recognition.}
We follow the experimental setting of \citep{atienza2021visiontransformerfastefficient}, training on MJSynth~\cite{Jaderberg14c} dataset and evaluating on IIIT5K~\cite{MishraBMVC12} dataset. 
Specifically, we first up-sample the low-resolution feature maps to high-resolution ones ($i.e.$, 48$\times$48) by using bilinear interpolation and a pre-trained VDIM. The resulting feature maps are then fed to a sequence encoder and a CTC head \cite{graves2006connectionist} to predict the class labels of characters in the images. 

We train the entire model with a global batch of 96 on 8$\times$A100s, using Adadelta optimizer with 5$e^{-2}$ learning rate, and a cosine scheduler for 6800 steps. We use the character-level cross-entropy loss between the predicted labels and the ground truth for supervision.
For evaluation, we report the sentence-level recognition accuracy of the test split of IIIT5K~\cite{MishraBMVC12} dataset in Fig.~\ref{fig:cv_jbu_bi_deconv}. 

\smallskip
\textbf{Linear probing semantic segmentation}. 
We follow the experimental setting of previous research \cite{fu2024featup, probe, STEGO}.
To be specific, on the COCOStuff dataset~\cite{lin2014mscoco}, we train a linear projection upon a frozen CLIP-ViT to directly predict the class label of each pixel. 

The input of the linear projection is the low-resolution feature maps. 
We train the linear projection with a global batch of 1024 on 8$\times$A100s, using Adam optimizer with 5$e^{-3}$ learning rate for 360 steps. We use the pixel-level cross-entropy loss between the predicted labels and the ground truth for supervision.
During the linear probing phase, we up-sample the low-resolution feature maps to a high-resolution ($i.e.$, 96$\times$96) one by using bilinear interpolation or pre-trained VDIM and then directly feed them into the linear projection to predict the pixel category. 
We report the segmentation accuracy on the validation split of COCOStuff in Fig.~\ref{fig:cv_jbu_bi_deconv}. 

\smallskip
\textbf{Fine-grained classification.}
We train and assess on CUB-200~\cite{keskar2016cub}, a fine-grained bird classification dataset. 
During the training, the low-resolution feature maps are first up-sampled to high-resolution ones (96$\times$96) with bilinear interpolation and a pre-trained VDIM. Then, a classification head with two linear layers pools the feature maps into a vector for image classification. Note that, the CLIP-ViT is frozen and only the parameters in the classification head are trainable.
We set the global batch as 16 on 1$\times$A100, using Adam optimizer with 1$e^{-4}$ learning rate and a cosine scheduler for 10 epochs.
We use the image-level cross-entropy loss between the predicted categories and the ground truth for supervision.
We report the image classification accuracy on the CUB-200 validation dataset in Fig.~\ref{fig:cv_jbu_bi_deconv}.

\begin{table}[tb]
  \setlength{\tabcolsep}{2.0pt}
  \caption{\color{black}Comparison of different choices of feature level on performance and efficiency. ESTVQA~\cite{wang2020estvqa} is a VQA benchmark focusing on scene text recognition.
}
  \centering
  \footnotesize
  \scalebox{0.9}{
  \begin{tabular}{c|cc|cccccccc}
    \toprule
      &\multicolumn{2}{c|}{Efficiency} & \multicolumn{7}{c}{Performance} \\
    \multirow{1}{*}{Level} &\rotatebox{90}{Period(h)} &\rotatebox{90}{Memory(G)} &\rotatebox{90}{Average} &\rotatebox{90}{GQA} &\rotatebox{90}{SQA} &\rotatebox{90}{REC} &\rotatebox{90}{VQA$^{\mathrm{C}}$} &\rotatebox{90}{VQA$^{\mathrm{T}}$} &\rotatebox{90}{ESTVQA} &\rotatebox{90}{MME$^{\mathrm{P}}$} \\ 
    \toprule
    \makecell[l]{\textit{0,2}} &27.7 &41.9 &63.4 &63.9 &69.5 &71.5 &60.5 &66.5 &40.6 &71.0 \\
    \makecell[l]{\textit{0,1,2}} &28.0 &41.9 &63.7 &63.8 &70.2 &71.8 &60.5 &66.9 &40.8 &72.1 \\
    \makecell[l]{\textit{0,1,2,3}} &45.6 &53.0 &63.8 &64.4 &69.3 &72.6 &60.7 &66.4 &41.6 &71.4 \\
    \makecell[l]{\textit{0,1,2,3}} (w/o MRL)  &45.6 &52.6 &62.4 &63.6 &69.8 &67.1 &57.8 &66.5 &39.9 &72.0 \\
    \toprule
  \end{tabular}
  }
\label{tab:features_level}
\end{table}

\begin{table}[t!]
\centering
\scalebox{0.74}{
\begin{tabular}{p{25mm} l| p{65mm}}
\toprule
Data & Size & Response formatting prompts \\
\midrule
LLaVA~\cite{liu2023llava1.5} & 158K & -- \\
ShareGPT~\cite{sharegpt} & 40K & -- \\
\midrule

VQAv2~\cite{goyal2017vqav2} & 83K & Answer the question using a single word or phrase. \\
GQA~\cite{hudson2019gqa} & 72K & \\
OKVQA~\cite{okvqa} & 9K & \\
OCRVQA~\cite{mishra2019ocrvqa} & 80K & \\
DocVQA~\cite{docvqa} & 15K & \\
ChartQA~\cite{chartqa} & 20K & \\
\midrule
A-OKVQA~\cite{schwenk2022okvqa} & 66K & Answer directly with the option's letter from the given choices. \\
\midrule

DVQA~\cite{kafle2018dvqa} & 20K  & -- \\
\midrule

TextCaps~\cite{sidorov2020textcaps} & 22K & Provide a one-sentence caption for the provided image. \\
\midrule

ShareGPT4V~\cite{sharegpt4v} & 55K & -- \\
\midrule

AI2D~\cite{Kembhavi2016ADI} & 3K & -- \\
\midrule

LAION-GPT4V~\cite{laion-gpt4v} & 11K & -- \\
\midrule

SythDog-EN~\cite{kim2022donut} & 40K & -- \\
\midrule

LRV-Instruct~\cite{liu2023aligning} & 30K & -- \\
\midrule

RefCOCO & 48K &  \\
\cite{kazemzadeh2014referitgame,mao2016generation} & & Provide a short description for this region. \emph{(for Region Caption)} \\
\cmidrule{1-2}
VG~\cite{krishna2017visual} & 86K & Provide the bounding box coordinate of the region this sentence describes. \emph{(for Referring Expression Comprehension)} \\
\midrule
Total & 858K & \\
\bottomrule
\end{tabular}
}
\caption{Detailed composition of our 858k-mixed dataset.}
\label{tab:sft_dataset}
\end{table}

\begin{table}[tb]
  \setlength{\tabcolsep}{2.0pt}
  \caption{Comparison of different choices of grid sizes on performance and efficiency. ``\textit{Pyramid}" means the feature grids from different levels form a region-level feature pyramid, $e.g.$, [2$\times$3] for level-0, [4$\times$6] for level-1, [8$\times$12] for level-2. ``\textit{Fix}" represents all feature maps are pooled into a 3$\times$3 feature grid. ``\textit{Selective}" represents adaptively selecting a grid size that is closest to the resolution. We measure the training period on 8$\times$A100s, the latency on an A100 with a 1008$\times$672 image, and the GPU memory on 8$\times$A100s with 1 image per GPU in supervised fine-tun ing phase.
}
  \centering
  \footnotesize
  {
  \begin{tabular}{c|ccc|c|cc|c|cccc}
    \toprule
      &\multicolumn{3}{c|}{Efficiency} & & \multicolumn{2}{c|}{General} & \multicolumn{1}{c|}{\makecell{Knowledge}} & \multicolumn{3}{c}{\makecell{OCR \& Chart}}  \\
    \multirow{1}{*}{Method} &Period(h) &Latency(s) &Memory(G) &Average &MME$^{\mathrm{P}}$ &GQA &AI2D &VQA$^{\mathrm{C}}$ &VQA$^{\mathrm{T}}$ &VQA$^{\mathrm{D}}$ \\
    \toprule
    \makecell[l]{\textit{Pyramid}} &62.4 &1.26 &60.3 &62.4 &69.0 &60.8 &57.3 &60.7 &67.5 &58.9 \\
    \makecell[l]{\textit{Fix [3$\times$3]}} &26.9 &0.62 &41.7 &64.6 &73.8 &63.9 &58.8 &60.9 &66.2 &63.8 \\
    \makecell[l]{\textit{Selective}} &27.7 &0.54 &39.4 &65.3 &73.0 &64.6 &58.3 &62.5 &68.5 &65.0 \\
    \toprule
  \end{tabular}
  }
\vspace{-0.3cm}
\label{tab:window_size}
\end{table}

\begin{figure*}[!t]
\centering
\includegraphics[width=1.0\linewidth]{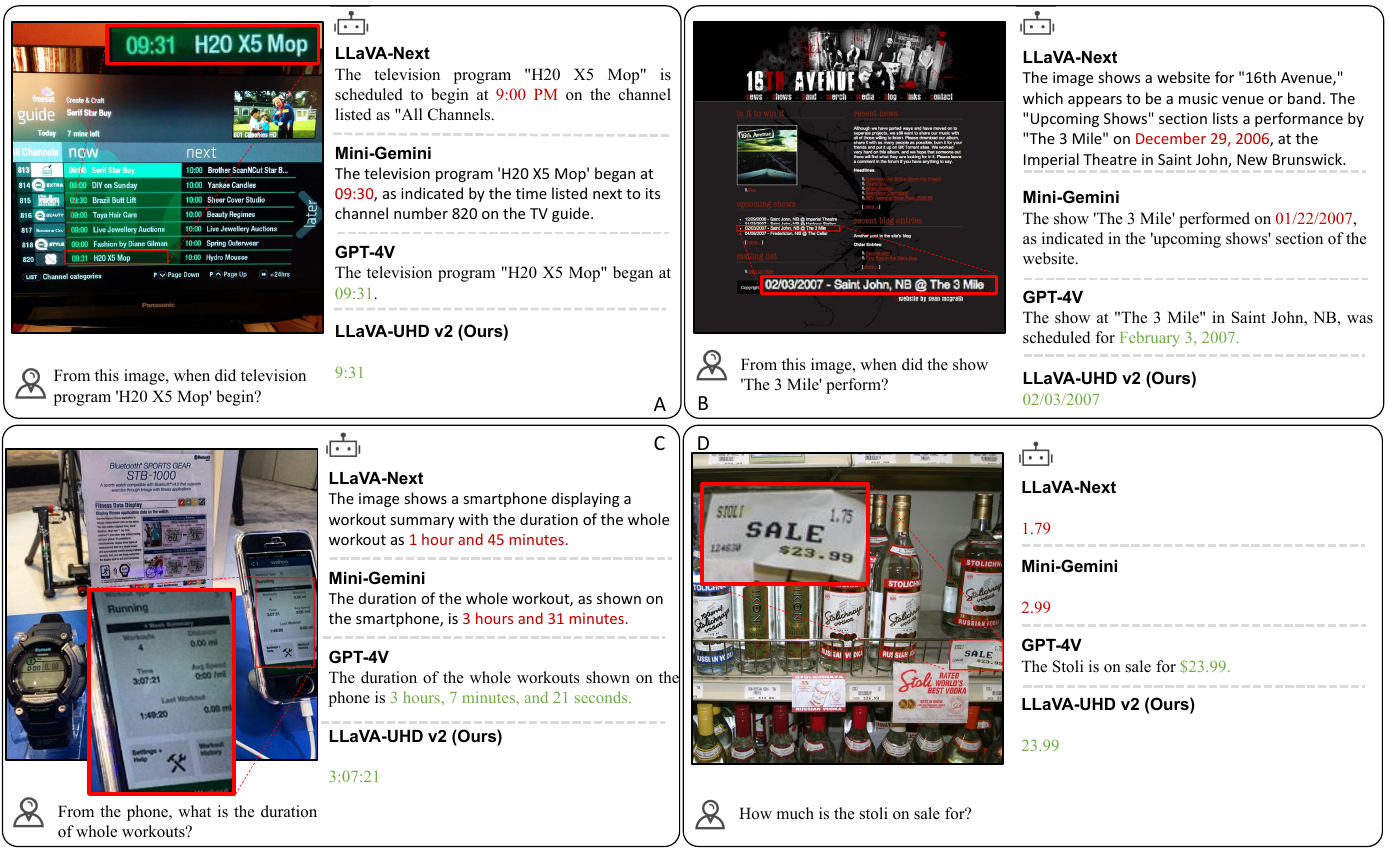}
\caption{Qualitative comparison on high-resolution dense perception task which requires the capabilities of fine-grained details perception.
}
\label{fig:case-dense}
\end{figure*}

\begin{figure*}[!t]
\centering
\includegraphics[width=1.0\linewidth]{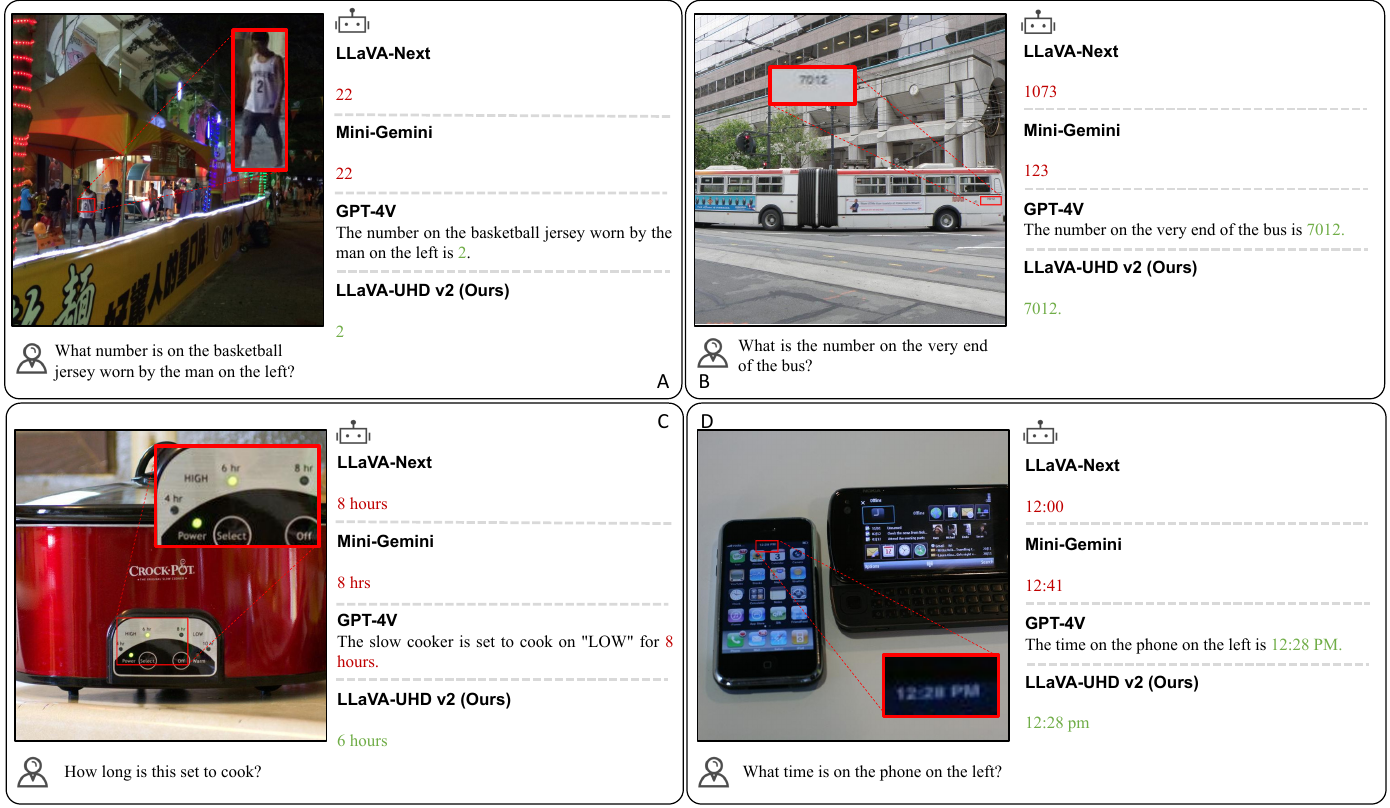}
\caption{Qualitative comparison on high-resolution fine-grained perception task which requires robust fine-grained visual texture perception capabilities.
}
\label{fig:case-small}
\end{figure*}

\begin{figure*}[!t]
\centering
\includegraphics[width=1.0\linewidth]{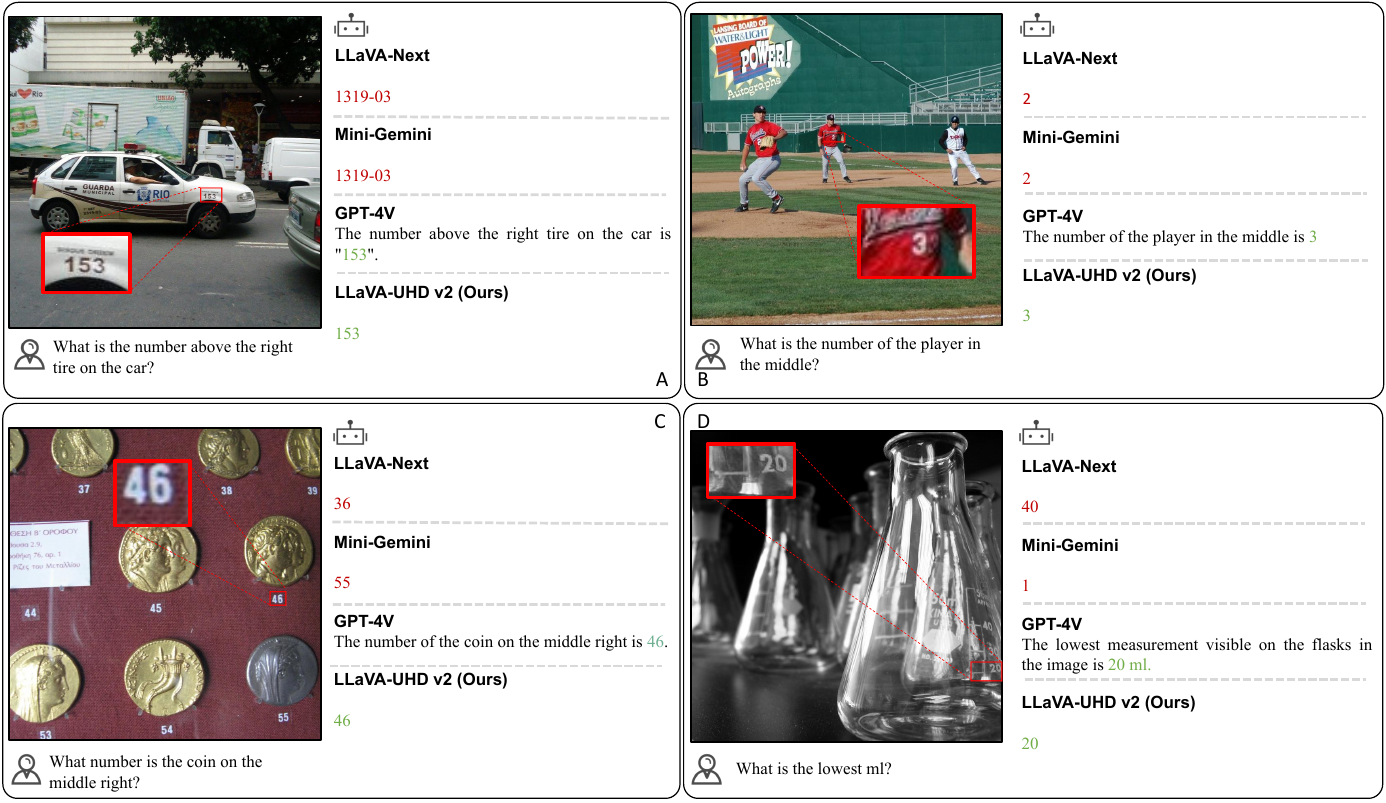}
\caption{Qualitative comparison on high-resolution spatial perception which necessitates the capabilities of high-level spatial contexts.
}
\label{fig:case-space}
\end{figure*}

\begin{figure*}[!t]
\centering
\includegraphics[width=1.0\linewidth]{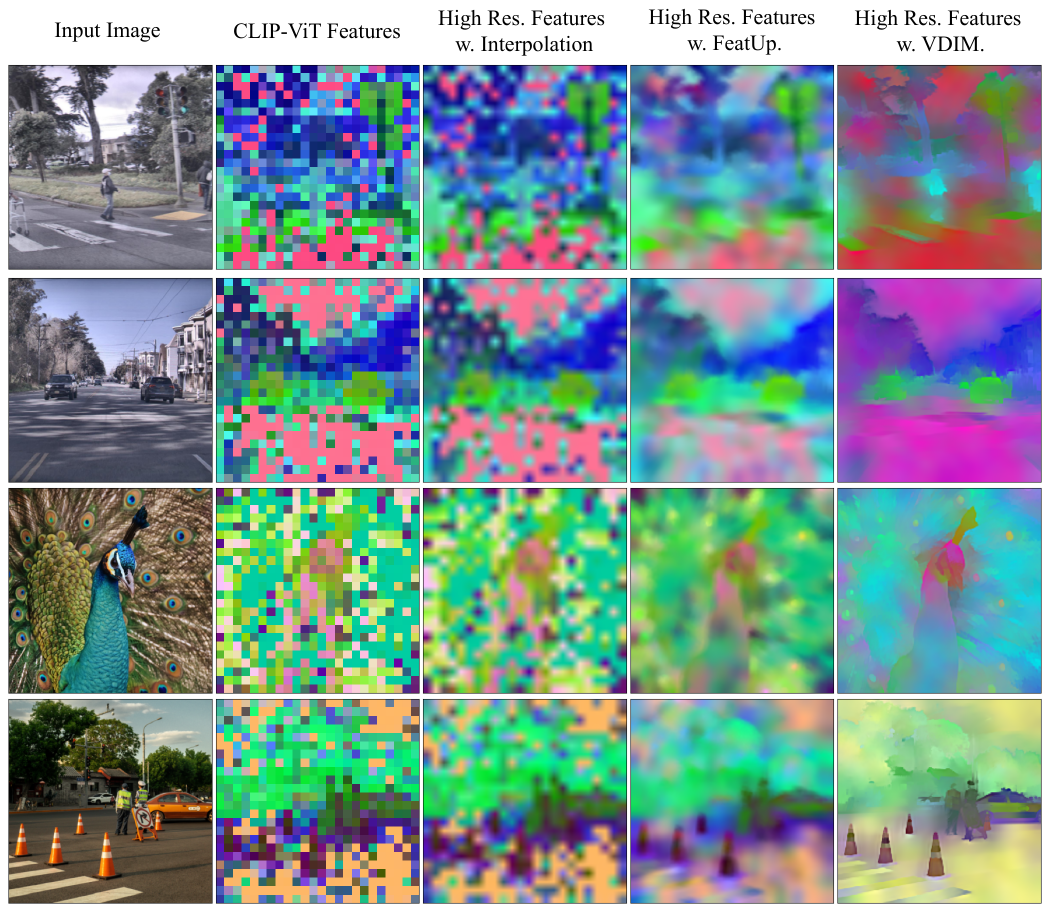}
\caption{PCA visualization of the up-sampled features by VDIM on nature scene. With VIDM, the high-resolution features could clearly depict object boundaries and text appearance. (Best viewed in color and zoomed in) }
\label{fig:jbu-visual1}
\end{figure*}

\begin{figure*}[!t]
\centering
\includegraphics[width=1.0\linewidth]{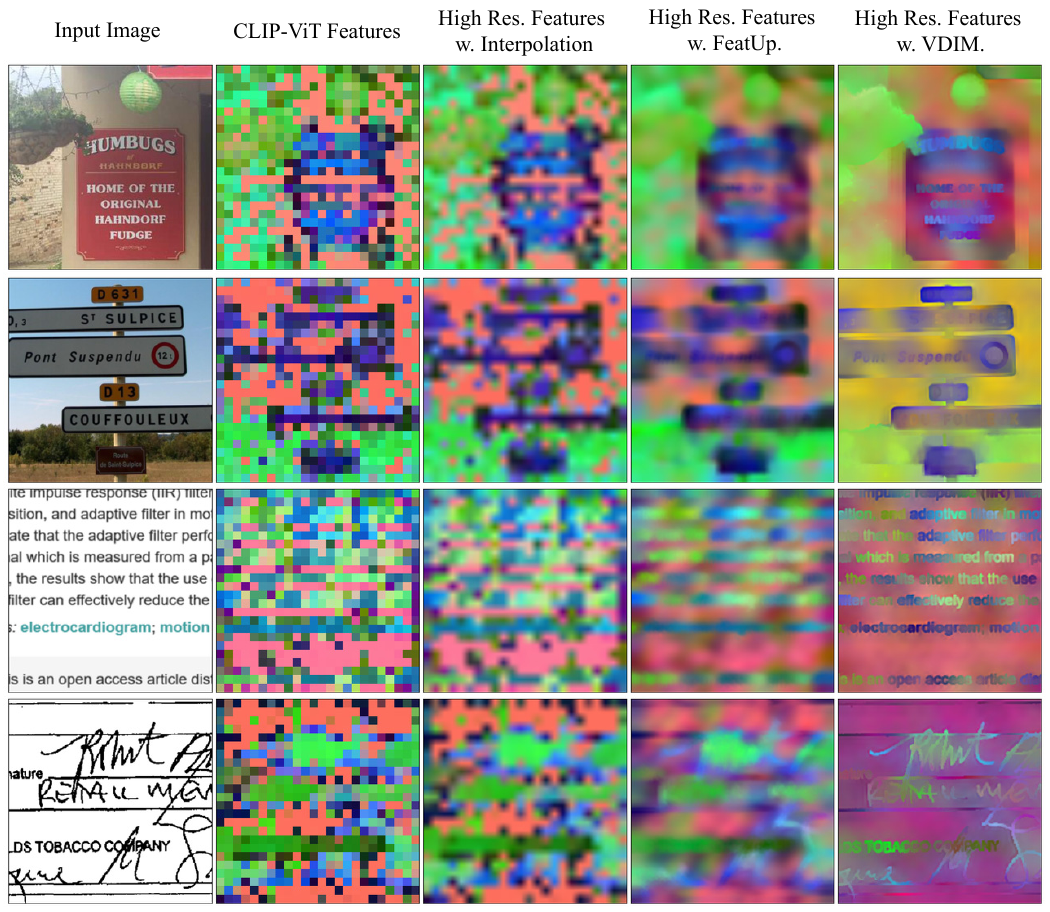}
\caption{PCA visualization of the up-sampled features by VDIM on OCR scene. With VDIM, the high-resolution features could clearly depict object boundaries and text appearance. (Best viewed in color and zoomed in) }
\label{fig:jbu-visual2}
\end{figure*}

\subsection{Visual detail injection module (VDIM)}
In this setction, we detail the implementation of the proposed VDIM.
Specifically, each pixel value of the up-sampled feature maps is extracted as 
\begin{equation}
    \label{eqn:in_conv}
    \resizebox{0.93\hsize}{!}{$
        \mathcal{F}^{l+1}[x,y] = \frac{1}{|U|} \sum_{(x',y') \in U} \biggl(\mathrm{Up}({\mathcal{F}^{l}})[x',y'] \times
        D_{dist} \times D_{sim}\big(\Theta^{l+1}(\mathcal{I}^{l+1}[x,y]),\Theta^{l+1}(\mathcal{I}^{l+1}[x',y'])\big)\biggr),
    $}
\end{equation}
where $[x,y]$ denotes the coordinate index of feature maps, $U$ the region of convolution neighborhood, $D_{dist}$ is a decay factor based on the distance between $[x,y]\ \text{and}\ [x', y']$, and $D_{sim}$ is a similarity weight based on the attention of each image pixel using a fully connected layer $\Theta^{l+1}$. 
By iteratively performing Eq.~\ref{eqn:in_conv}, we progressively up-sample the feature maps and finally construct the ISP $\{\mathcal{F}^0, \mathcal{F}^1, \mathcal{F}^2\}$.

\smallskip
\textbf{Up-sampling kernel}. Following \cite{hamilton2020likely}, the kernel weight $\Theta^{l+1}$ in Eq.\ref{eqn:in_conv} of the VDIM relies on two weights $D_{dist}$ and $D_{sim}$. 
Spatial distance decay $D_{dist}$ is defined as 
\begin{equation}
    \label{eqn:dist}
    \begin{split}
    D_{dist} = \mathrm{exp}\left(-\frac{\lVert(x,y)-(x',y')\lVert_2^2}{2\sigma_{dist}^2}\right)
    \end{split}
\end{equation}
to represent the Euclidean distance relations between adjacent pixels, where {\color{black}$\sigma_{dist}$} denotes a learnable width. 
And pixel similarity weight $D_{sim}$ is determined as 
\begin{equation}
    \label{eqn:sim}
    \begin{split}
    D_{sim} = \underset{(x',y')\in U}{\mathrm{softmax}}\left(\frac{\Theta^{l+1}(\mathcal{I}^{l+1}[x,y])\cdot\Theta^{l+1}(\mathcal{I}^{l+1}[x',y'])}{\sigma_{sim}^2}\right),
    \end{split}
\end{equation}
where {\color{black}$\sigma_{sim}$} is a learnable temperature factor to modulate the distribution of similarity scores and $U=7$.

\smallskip
\textbf{Learnable down-sampler.} We detail the implementation of the learnable down-sampler defined in Eq.2 of the main paper. Compared to a simple convolutional layer with a stride of 2, we apply an attention down-sampler following ~\cite{fu2024featup} at each feature level. 

Specifically, a 1$\times$1 convolution layer is first carried out on the feature maps of $(l+1)$-th level to extract a saliency map, followed by combining it with a modified fully-connected layer to normalize the features in the local neighborhood $V$. We summarize the above operation as a network $f(\cdot)$ with trainable parameters $\Omega^{l+1}$.
As a result, the feature pixel at the location of $[x,y]$ on down-sampled feature maps (of $l$-th level) is formally defined as
{ 
\begin{equation}
    \resizebox{0.8\hsize}{!}{$
    \label{eqn:down}
    \mathrm{Down}(\mathcal{F}^{l+1};\Omega^{l+1})[{x,y}] = \\ \text{softmax}\left(f(\mathcal{F}^{l+1}[V_{x,y}];\Omega^{l+1})\right) \cdot \mathcal{F}^{l+1}[V_{x,y}],
    $}
\end{equation}
}
where $V_{x,y}$ denotes the local neighborhood in the high-resolution feature maps. {Note that, before performing Eq.~\ref{eqn:down}, we experimentally up-sample the $\mathcal{F}^{l+1}$ to the size of the original image using bilinear interpolation. And we set $V_{x,y}=14$}, aligning with the patch size of CLIP-ViT, to simulate the feature extraction of ViT.

\subsection{Supervised fine-tuning dataset}

As illustrated in Table~\ref{tab:sft_dataset}, we detail the proposed 858k-mixed dataset in the supervised fine-tuning phase of MLLM.

\section{Analysis}
\label{sec:Analysis}

\subsection{Quantitative experiment}

\textbf{Level choice.}
As shown in Table~\ref{tab:features_level}, the introduction of higher level (higher resolution) feature maps results in consistent enhancement of the average performance. However, incorporation of even higher resolution features, such as level-3 feature maps ($i.e.$, $8\times$ resolution than level-0), yields marginal benefits while substantially increasing the training cost. 

When the multi-level reconstruction (MLR) supervision is replaced by the last-level supervision, performance deteriorates. In such a scenario, the VDIM faces substantial challenges in effectively incorporating high-frequency information into CLIP-ViT features.

\textbf{The choice of RoI-align grid sizes} 
We explore the impact of RoI-align grid size on the efficiency and performance of LLaVA-UHD v2. Specifically, we RoI-align a region-level feature pyramid from the inverse feature pyramid within a window set, ensuring that higher-resolution feature maps retain finer-grained pooling grids, like~\cite{tong2024cambrian}. However, this approach, rather than improving multi-scale feature integration, significantly degrades performance and inference efficiency, as demonstrated in Table \ref{tab:window_size}.
Compared to fixed grid, selecting a proper pooling grid (defined in Eq.3 of main paper) showcases better performance and efficiency, because of a more approximate aspect ratio to the native image.

\subsection{Qualitative experiment}

\smallskip
\subsubsection{Enhanced high-resolution features.} 
We performed a qualitative visualization of our ISP, presented in Fig.\ref{fig:jbu-visual1} and Fig.\ref{fig:jbu-visual2}, Note that all the high-resolution features shown are the highest-level of ISP enhanced from CILP-ViT features. While bilinear interpolation increases the nominal resolution of features, it fails to enhance the fidelity of image detail representation. Using the FeatUp~\cite{fu2024featup} for feature up-sampling, the resulting features capture finer details but retain a degree of blurriness.
In contrast, with the proposed VIDM, the high-resolution features clearly
depict object boundary and text appearance, demonstrating accurate and refined representation of visual details.

\subsubsection{Case studies.} 
\label{appendix:cases}

We add more cases to analyze the behavior of our LLaVA-UHD v2. The capabilities are summarized as three aspects as follows.

\smallskip

\textbf{Dense Perception.}
{In Fig.~\ref{fig:case-dense}, we visualize the performance of well-known MLLMs on dense perception tasks. Dense perception tasks require models to possess highly robust fine-grained perception capabilities to distinguish object boundaries within a large number of densely packed similar objects, thereby accurately locating the target and its boundaries to identify the target precisely.

It is evident that LLaVA-UHD v2 and GPT-4V accurately identify the beginning time of the television program `H20 X5 Mop' (case A), the performance date of the show `The 3 Mile' (case B), the duration of whole workouts (case C), and the prize of Stoli (case D), indicating highly robust fine-grained perception capabilities provided by our visual pyramid representation. In comparison, other models either fail to precisely locate the target (LLaVA-Next) or cannot distinguish the target from similar adjacent objects, limited in accurately completing dense OCR tasks (Mini-Gemini).}

\textbf{Fine-grained Perception.}
{In Fig.~\ref{fig:case-small}, we visualized the performance of well-known MLLM on fine-grained perception tasks. These tasks require models to have robust fine-grained perception capabilities to detect the textures of small or blurry targets, thereby accurately locating and identifying small targets. 

Case C indicates that LLaVA-UHD v2 accurately identified the small green light, and the tiny number of duration time associated with green light, demonstrating that the introduction of high-frequency information in hierarchical features can handle small, blurry targets effectively. In contrast, other models  can not find the small green light, or fail to accurately perform OCR tasks due to the text being too small or blurry (\textit{e.g}., GPT-4V, LLaVA-Next, Mini-Gemini). 

This capability is further demonstrated in cases A, B and D, where both LLaVA-UHD v2 and GPT-4V accurately identified the tiny number on the basketball jersey (case A), the blurry number on the very end of the bus (case B), and the time on the phone (case D), while LLaVA-Next and Mini-Gemini exhibited limitations.}

\textbf{Spatial Perception.}
{In Fig.~\ref{fig:case-space}, we visualized the performance of well-known MLLM on spatial perception tasks. Spatial perception tasks require models to have robust high-semantic perception capabilities to discern the spatial relationships between different objects. 

It is evident that LLaVA-UHD v2 and GPT4V perceive the spatial relative positions between different objects, to accurately identify the number above the right tire on the car (case A), the number of the player in the middle (case B), the number of the coin on the middle right (case C), the lowest ml (case D). This accuracy is attributed to our high-resolution visual pyramid representation, which allows the perfect integration of features of varying semantics and spatial information across different levels. In contrast, other models, such as LLaVA-Next and Mini-Gemini, fail to accurately perceive these relative spatial positions. }

\end{document}